\documentclass[letterpaper, 10 pt, conference]{ieeeconf}  

\usepackage{graphicx}
\usepackage{subfigure}

\IEEEoverridecommandlockouts                              
\title{\LARGE \bf
 The Field-of-View Constraint of Markers for Mobile Robot with Pan-Tilt Camera
}

\author{Hongxuan Ma,  Wei Zou, Zheng Zhu, Siyang Sun, Zhaobing Kang
\thanks{*This work is supported by The National Key Research and Development Program of China (Project 2018YFB1306303), and The National Natural Science Foundation of China (Grant No. 61773374).}
\thanks{Hongxuan Ma, Zheng Zhu, Siyang Sun and Zhaobing Kang are affiliated to Institute of Automation, Chinese Academy of Science, University of Chinese Academy of Sciences; mahongxuan2016@ia.ac.cn
}%
\thanks{Wei Zou is affiliated to Institute of Automation, Chinese Academy of Science, Beijing China; wei.zou@ia.ac.cn}%
}

\begin{document}

\maketitle
\thispagestyle{empty}
\pagestyle{empty}

\begin{abstract}
 In the field of navigation and visual servo, it is common to calculate relative pose by feature points on markers, so keeping markers in camera's view is an important problem. In this paper, we propose a novel approach to calculate field-of-view ($FOV$) constraint of markers for camera. Our method can make the camera  maintain the visibility of all feature points during the  motion of mobile robot. According to the angular aperture of camera, the mobile robot can obtain the $FOV$ constraint region where the camera cannot keep all feature points in an image. Based on the $FOV$ constraint region, the mobile robot can be guided to move from the initial position to destination. Finally simulations and experiments are conducted based on a mobile robot equipped with a pan-tilt camera, which validates the effectiveness of the method to obtain the $FOV$ constraints.

\end{abstract}

\vspace{-2mm}
\section{Introduction}

In the field of navigation, many robots locate themselves by other objects, so keeping them in camera's view is a important  problem. \cite{Azouaoui2014Passively,Bouraine2012Provably,Laurent2013Keeping,Lopez2007Switched,kang2019adaptive,shao2013motion,choi2019deep}. In the field of servoing, it is also necessary to keep objects in the camera's view \cite{Fang2012Adaptive,stolkin2008efficient,lopez2017formation,Park2012Novel,zhu2019scalable}.
However, for the camera, there are some $FOV$ constraint regions where the camera cannot keep objects in a view.
%
The solutions to such $FOV$ problem for mobile robot can be divided into two categories: tracking based on the pan-tilt camera platform and path planning based on $FOV$ constraint.

 There are many researches about the mobile robot equipped with a pan-tilt camera. Fang \emph{et al.} \cite{Fang2012Adaptive} design an adaptive pan camera controller to track the feature points. Fan \emph{et al.} \cite{Fan2018Robust} mount a pan-tilt camera to track the target when the mobile robot moves. Le and Pham \cite{Le2016Optimal} present a dynamic control method to track a moving object by a mobile robot with pan-tilt binocular camera. However, there are some positions that are too close to these points, and it is impossible to keep all points in the camera's view.

Another common method is by path planning based on the $FOV$ constraint. Bhattacharya \emph{et al.} \cite{Bhattacharya2005Path} present a control scheme to provide the minimal length path for a mobile robot, which can maintain visibility of a landmark. Then they study methods to maintain the visibility of a fixed landmark for a mobile robot with limited sensing in \cite{Bhattacharya2006Controllability,Bhattacharya2007Optimal}. Based on the extremal paths that saturate the camera pan angle, the optimal path is given.
Salaris \emph{et al.} \cite{Salaris2010Shortest} propose a complete characterization of the shortest paths to the desired position. A global partition of motion plane is presented based on the limited $FOV$ in left and right direction, and according to the initial position the mobile robot moves along different paths to the desired position. After the paper \cite{Salaris2010Shortest}, Salaris \emph{et al.} \cite{Salaris2013Shortest} introduce the vertical constraint to the $FOV$ problem. Later Salaris \emph{et al.} further study the $FOV$ constraints of limited camera, and propose control method to drive the mobile robot to a desired position with a shortest path\cite{Salaris2011From,Salaris2015Epsilon}.


\vspace{-2mm}
\begin{table}[h]
\caption{ }
\centering
\begin{tabular}{|l|l|}
    \hline
       \quad Shorthand &     \quad\quad\quad\quad\quad\quad\quad\quad Meaning\\
    \hline
           \quad $RNH$  &     The Region where the pan-tilt camera can Not \\&observe all feature points in the Horizontal direction.\\
    \hline
            \quad $RNV$  &    The Region where the pan-tilt camera can Not \\&observe all feature points in the Vertical direction.\\
    \hline
            \quad $BH$  &     A rectangular Box used to calculate $RNH$.\\
    \hline
             \quad $BV$  &     A rectangular Box used to calculate $RNV$.\\
    \hline
            \quad $RA$  &     The Region where the pan-tilt camera can \\&keep All feature points in a view.\\
    \hline
            \quad $RNA$ &     The Region where the pan-tilt camera can Not \\&keep All feature points in a view. It is the union \\&of $RNH$ and $RNV$. \\
    \hline
\end{tabular}
\\
\label{Table1}
\end{table}
\vspace{-2mm}
When features points are separated with large distance or the camera is too close to them, the camera is unable to keep all feature points in the view even though it is mounted on a pan-tilt platform. Many methods to obtain the $FOV$ constraint by regarding marker as a point. To maintain visibility, some redundant paths are always needed.
In this paper, we present an approach to maintain the visibility of all feature points on markers by calculating $FOV$ constraint for a mobile robot with pan-tilt camera. For convenience, some abbreviations in Table \ref{Table1} are used for description. Based on the angular aperture of camera, the mobile robot can calculate two kinds of boxes $BH$ and $BV$, then $RNH$ and $RNV$ can be obtained from $BH$ and $BV$, respectively. According to $RNH$ and $RNV$, the $FOV$ constraint regions where the mobile robot cannot keep markers in the view can be acquired. Compared with the methods mentioned above, our approach has such advantages: 1) it can be applied to observing points on multiple markers. 2) the $FOV$ constraint regions are calculated explicitly, which can guide the mobile robot to avoid regions where the mobile robot cannot keep markers in the view. 3) it combines the advantage of tracking based on pan-tilt camera and path planning based on $FOV$ constraint, so it has less redundant paths.

The remainder of this paper is organized as follows. In section II, some related concepts are introduced. Section III describes the method to calculate $FOV$ constraint region $RNA$. In section IV, the results of simulations and experiments are provided. In section V, the conclusion is given.

\section{Preliminaries}

\subsection{The $FOV$ Model of Camera}
\begin{figure}[htbp]
\small
\centering
\includegraphics[width=4cm]{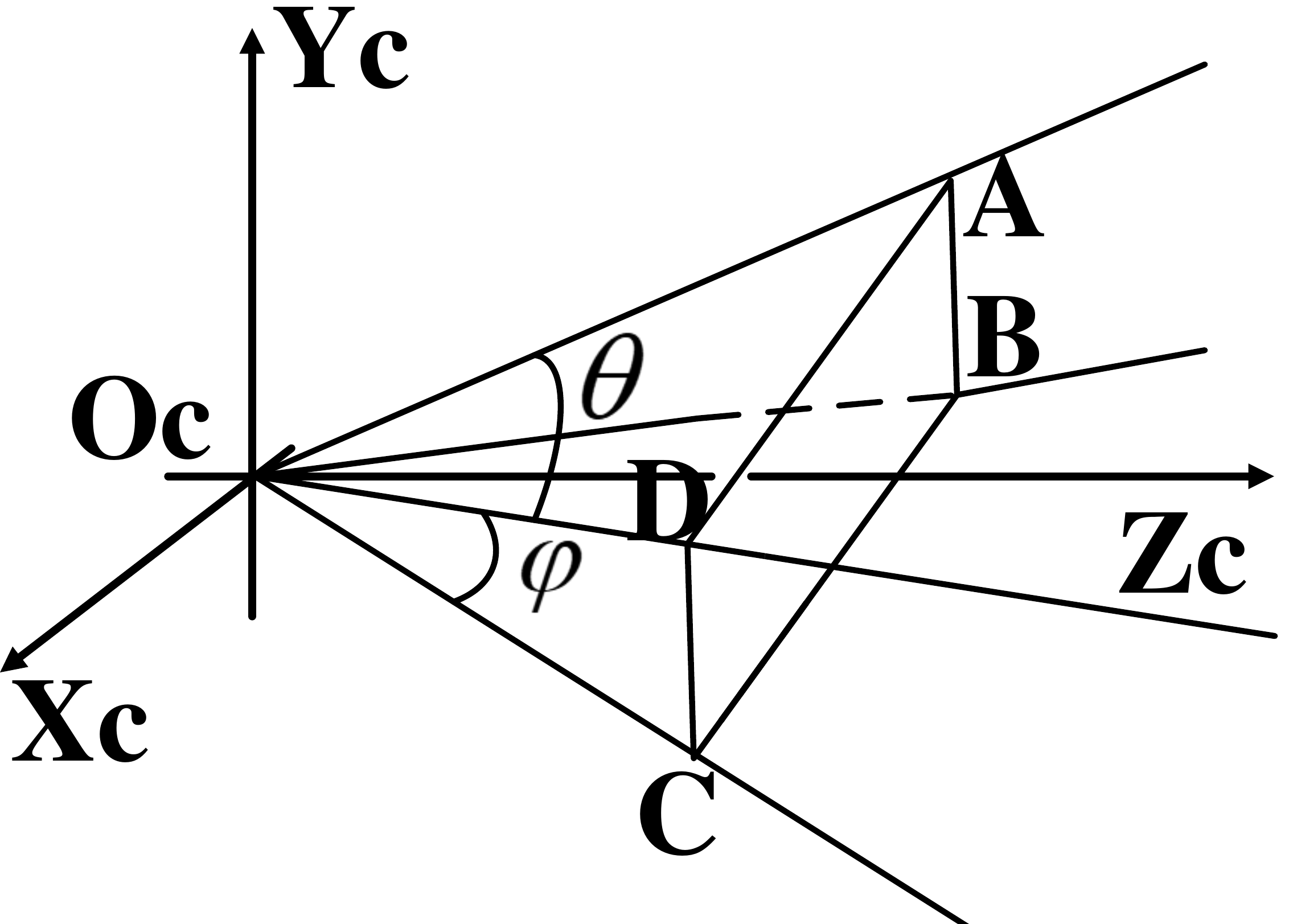}\\
\caption{The $FOV$ Model of Camera: the rectangle $ABCD$ is the image plane.}
\label{angleView}
\end{figure}

In Fig. \ref{angleView} the view field of the camera can be modeled as a four-side right rectangular pyramid, and only the points inside the pyramid can be observed by the camera. In this figure, $O_c$ is the optical center of the camera, and $\theta$ is the Horizontal-FOV angular aperture and $\varphi$ is the Vertical-FOV angular aperture.

Among these points, the left-most, right-most, top-most, bottom-most points are denoted as $(u_l,v_l)$, $(u_r,v_r)$, $(u_t,v_t)$ and $(u_b,v_b)$ with $u_l=min\{u_i, i=1,...,N\}$, $u_r=max\{u_i, i=1,...,N\}$, $v_t=min\{v_i, i=1,...,N\}$ and $v_b=max\{v_i, i=1,...,N\}$, respectively.
In the mobile robot coordinate system, their coordinates are represented by $P_l$, $P_r$, $P_t$, $P_b$, respectively. For a $W \times H$ image, when $0<u_l$ and $u_r<W$, it can be considered that the all points are kept in the view in horizontal direction. When $0<v_t$ and $v_b<H$, all points can be considered to be kept in the view in vertical direction.


\section{The Method to Calculate $RNH$ and $RNV$}

In our method, $RNA$ is the union of $RNH$ and $RNV$, $RNH$ and $RNV$ are obtained from $BH$ and $BV$ respectively, so it is necessary to calculate the $BH$ and $BV$ at first. $BH$ and $BV$ are calculated based on the angular aperture of camera and some geometric relations. In this section the following coordinates are all denoted in mobile robot coordinate system except for special instructions.
%
%
%

\subsection{The Calculation of $BH$ and $BV$}
    By Perspective-n-Point ($PnP$) method, the coordinate of feature points in camera coordinate system can be calculated. Through coordinate transformation, their coordinates in mobile robot system can also be obtained.

\begin{figure}
  \subfigure[]{
    \includegraphics[width=1.1in]{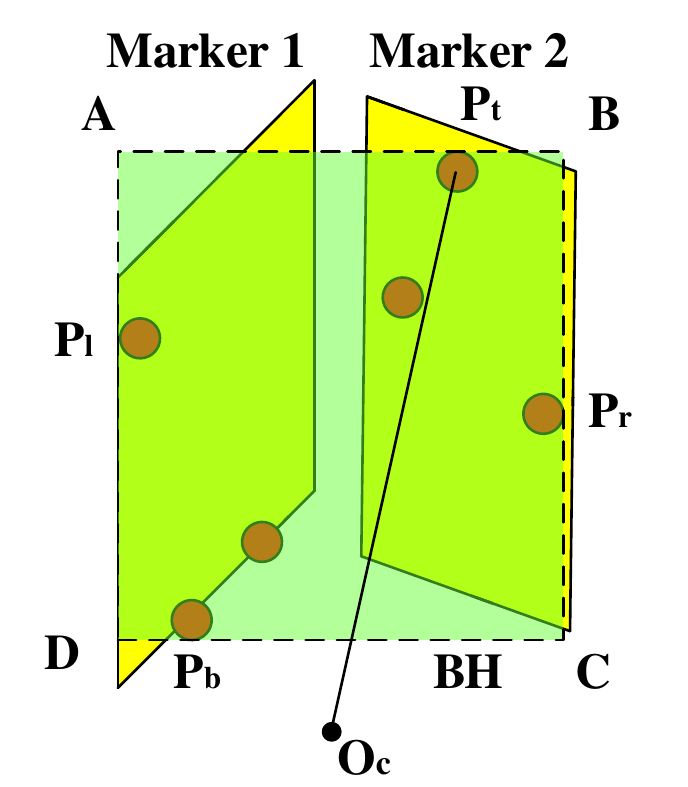}
    \label{drawHorizontalConstraintBox}
    }
    \setlength{\belowdisplayskip}{3pt}
    \setlength{\belowdisplayskip}{3pt}
    \setlength{\belowdisplayskip}{3pt}
    \setlength{\belowdisplayskip}{3pt}
    \setlength{\belowdisplayskip}{3pt}
    \setlength{\belowdisplayskip}{3pt}
    \setlength{\belowdisplayskip}{3pt}
    \setlength{\belowdisplayskip}{3pt}
    \setlength{\belowdisplayskip}{3pt}
    \setlength{\belowdisplayskip}{3pt}
  \subfigure[]{
    \includegraphics[width=1.1in]{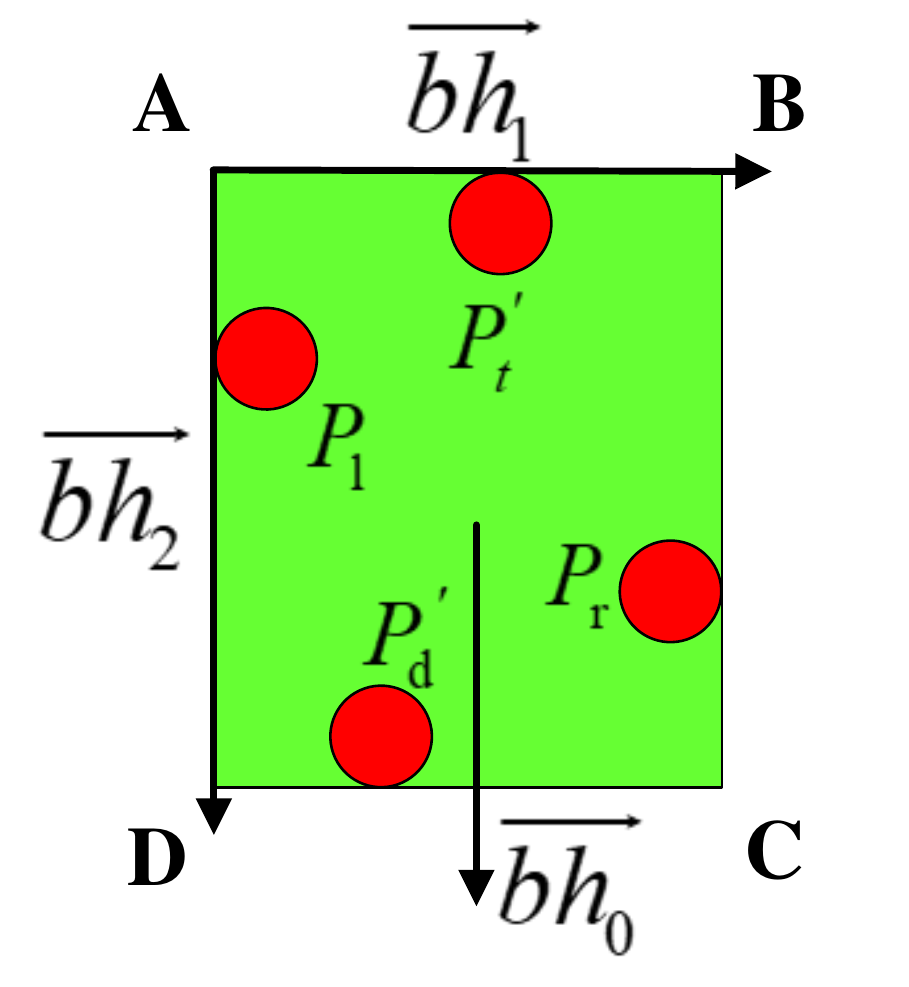}
    \label{HorizontalConstraintBox}
    }
    \caption{(a) The schematic to calculate $BH$, the points on Marker 1 and Marker 2 are used to calculate the relative pose. (b) $BH$.}
\end{figure}

     (1) Determination of $BH$: To acquire the vertex coordinates of $BH$, three vectors need to be calculated at the beginning. $\overrightarrow {{bh_0}} (x_0,y_0,z_0)$ is the normal vector to $BH$, and it is the average of unit normal vectors of Marker 1 and Marker 2 in Fig. \ref{drawHorizontalConstraintBox}.
     $\overrightarrow {{bh_1}} (x_1,y_1,0)$ is a unit vector in the horizontal direction, which is perpendicular to $\overrightarrow {{bh_0}}$. $\overrightarrow {{bh_2}} (x_2,y_2,z_2)$ is a unit vector perpendicular to $\overrightarrow {{bh_0}} $ and $\overrightarrow {{bh_1}}$.

      Based on equation (\ref{1})-(\ref{3}), the three unit vectors can be obtained.
        \begin{equation}
        \overrightarrow {{bh_0}}   \cdot \overrightarrow {{bh_1}}   = 0\label{1}
        \end{equation}
        \begin{equation}
        \overrightarrow {{bh_0}}   \times \overrightarrow {{bh_1}}=\overrightarrow {{bh_2}} \label{2}
         \end{equation}
        \begin{equation}
        \left| {\overrightarrow {{bh_1}}} \right| = \left| {\overrightarrow {{bh_2}}} \right| = 1 \label{3}
        \end{equation}

\vspace{-1mm}
        In Fig. \ref{drawHorizontalConstraintBox}, $P_t$, $P_b$, $P_l$ and $P_r$ can be obtained by $PnP$ method, and the two non-horizonal side of $BH$ pass the $P_l$ and $P_r$, respectively. The vector from $P_{t}$ to $O_c$ is $\overrightarrow {{bh}}$. The projection of $P_t$ on $BH$ along $\overrightarrow {{bh}}$ is denoted as $P_{t}^\prime$. The coordinate of $P_{t}^\prime$ can be obtained by (\ref{5}) and (\ref{4}).
       \begin{equation}
       \overrightarrow {{P_{t}}^\prime{P_l}}\cdot \overrightarrow {{bh_0}} =0  \label{5}
       \end{equation}
        \begin{equation}
        {P_t} + m_1 \cdot \overrightarrow {{bh}}   = {P_{t}^\prime} \label{4}
        \end{equation}
where $m_1$ is a scalar coefficient, and it also can be obtained from (\ref{5}) and (\ref{4}).
 Similarity, the projection of $P_b$ on $BH$ can be obtained, and it is denoted as $P_{b}^\prime$.
According to $P_l$, $P_r$, $P_{t}^\prime$ and $P_{b}^\prime$, the four vertices (denoted as $A, B, C, D$) of $BH$ can be obtained. Because two sides of $BH$ pass through points $P_l$ and $P_r$ respectively, the coordinate of point $A$ in Fig. \ref{HorizontalConstraintBox} can be calculated by (\ref{6}) and (\ref{7}).
\begin{equation}
{P_l} + p_1 \cdot \overrightarrow {{bh_2}}   = {P_{t}^\prime} + q_1 \cdot \overrightarrow {{bh_1}}  \label{6}
\end{equation}
\begin{equation}
A = {P_l} + p_1 \cdot \overrightarrow {{bh_2}}       \label{7}
\end{equation}
where $p_1$ and $q_1$ can also be calculated from Equ. (\ref{6}), (\ref{7}).
Similarity, B, C, D of $BH$ can be obtained. $BH$ can be used to calculate the horizontal $FOV$ constrain.

(2) Determination of $BV$:
$BV$ is a rectangular box, and its two horizonal sides pass $P_t$ and $P_b$, respectively. To acquire the vertex coordinates of $BV$, three vectors also need to be calculated at the
beginning. According to the coordinates of feature points, the average unit normal vector $\overrightarrow {{bv_0}}$ of Marker 1 and Marker 2 can be obtained. $\overrightarrow {{bv_1}}$ is a vector in the horizontal direction, which is perpendicular to $\overrightarrow {{bv_0}}$. $\overrightarrow {{bv_2}}$ is a vector perpendicular to $\overrightarrow {{bv_0}}$ and $\overrightarrow {{bv_1}}$. The three vectors can be obtained by (\ref{8})-(\ref{10}).

    \begin{equation}
    \overrightarrow {{bv_0}}  \cdot \overrightarrow {{bv_1}}  = 0\label{8}
    \end{equation}
    \begin{equation}
    \overrightarrow {{bv_0}}  \times   \overrightarrow {{bv_1}}=\overrightarrow {{bv_2}}  \label{9}
     \end{equation}
    \begin{equation}
    \left| {\overrightarrow {{bv_1}}} \right| = \left| {\overrightarrow {{bv_2}}} \right| = 1 \label{10}
    \end{equation}

        The projection of $P_l$ on $BV$ is denoted as $P_{l}^\prime$. The vector from $P_{l}$ to $O_c$ is $\overrightarrow {{bv}}$. The coordinate of $P_{l}^\prime$ can be obtained by equation (\ref{111}) and (\ref{11}).

\begin{equation}
\overline {{P_l}^\prime {P_t}}  \cdot \overrightarrow {{bv_0}}=0\label{111}
\end{equation}
\begin{equation}
{P_l} + m_2 \cdot \overrightarrow {{bv}}  = {P_{l}^\prime} \label{11}
\end{equation}
where $m_2$ is a scalar coefficient, and it can also be obtained from (\ref{111}) and (\ref{11}).
       Similarity, $P_{r}^\prime$ can be obtained. The coordinate of $A$ can be obtained by (\ref{12}) and (\ref{13}).
        \begin{equation}
        {P_{l}^\prime} + p_2 \cdot \overrightarrow {{bv_2}}   = {P_t} + q_2 \cdot \overrightarrow {{bv_1}} \label{12}
        \end{equation}
        \begin{equation}
        A = {P_{t}} + q_2 \cdot \overrightarrow {{bv_1}} \label{13}
        \end{equation}

Similarity, based on $P_t$, $P_b$, $P_{l}^\prime$, and $P_{r}^\prime$, the coordinates of B, C, D of $BV$ can be calculated. Based on $BV$, the vertical $FOV$ constrain can be obtained.

\subsection{Horizontal $FOV$ Constrain}

\begin{figure}[htbp]
\small
\centering
\includegraphics[width=6cm]{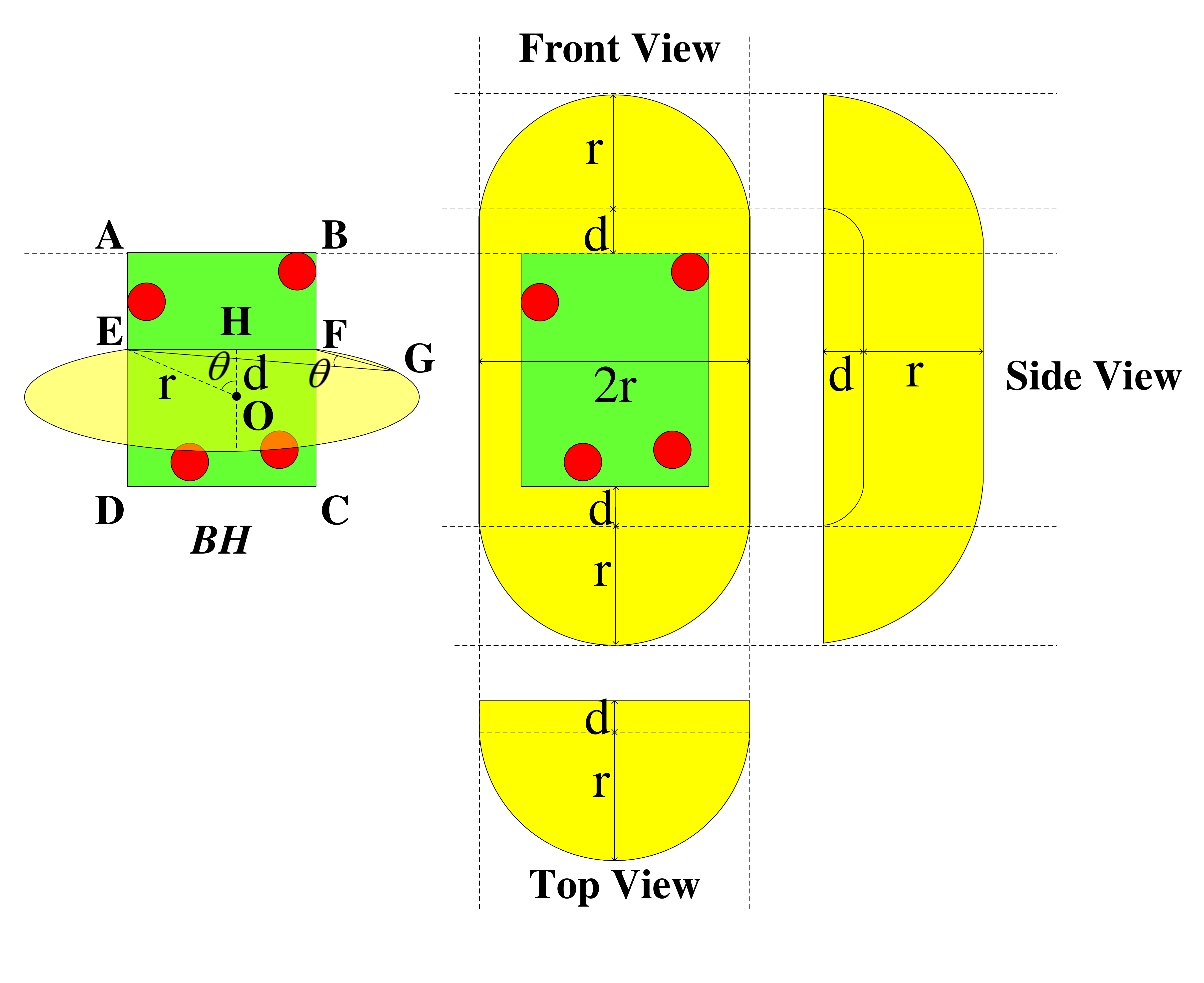}\\
\caption{The left part of the figure is the schematic to calculate $Sh_{ABCD}$. The right part are the three views of $Sh_{ABCD}$.}
\label{horizontalConstrains}
\end{figure}

Based on $BH$, the 3D space with horizontal $FOV$ constraint can be obtained. In this space the camera cannot keep all feature points in a view in the horizontal direction. As Fig. \ref{horizontalConstrains} shown, $Plane$ $EOF$ is perpendicular to $BH$. $\overline{EF}$ is a line segment parallel to $\overline{AB}$ and it intersects lines $\overline{AD}$ and $\overline{BC}$ at point $E$ and point $F$, respectively. $\theta$ is the Horizontal-FOV angular aperture of camera. For a camera (denoted as $G$) on $Plane$ $EOF$, if points $E$ and $F$ are just kept in its view in the horizontal direction, its trajectory is part of a circle, which can be proven based on $The$ $Center$ $Angle$ $Theorem$.
The circle $EFG$ (the yellow region) rotates $\pm \pi/2$ around axis $\overline{EF}$, then all the covered space ($Sh_{EF}$) can be obtained. The positions outside of $Sh_{EF}$ can make the camera keep points $E$ and $F$ in its view in the horizontal direction. Along segment $\overline{AD}$, there are infinite $\overline{EF}$ and the corresponding $Sh_{EF}$, the union of all $Sh_{EF}$ is $Sh_{ABCD}$. Outside of $Sh_{ABCD}$ the camera can keep segments $\overline{AD}$ and $\overline{BC}$ in its view in the horizontal direction, otherwise it cannot keep them in its view in the horizontal direction. In Fig. \ref{horizontalConstrains}, the length of $\overline{EF}$ is $l_{EF}$. $r$ and $d$ can be obtained by Equ(\ref{14})-(\ref{15}). They can be calculated by:

\begin{equation}
{l_{EF}}=\sqrt {{{({x_A} - {x_B})}^2} + {{({y_A} - {y_B})}^2} + {{({z_A} - {z_B})}^2}} \label{14}
\end{equation}
\vspace{-2mm}
\begin{equation}
r = \frac{l_{EF}}{{2\sin \theta }} \label{15}
\end{equation}
\vspace{-2mm}
\begin{equation}
d = \frac{l_{EF}}{{2\tan \theta }} \label{16}
\end{equation}

\begin{figure}
  \subfigure[]{
    \includegraphics[width=1.5in]{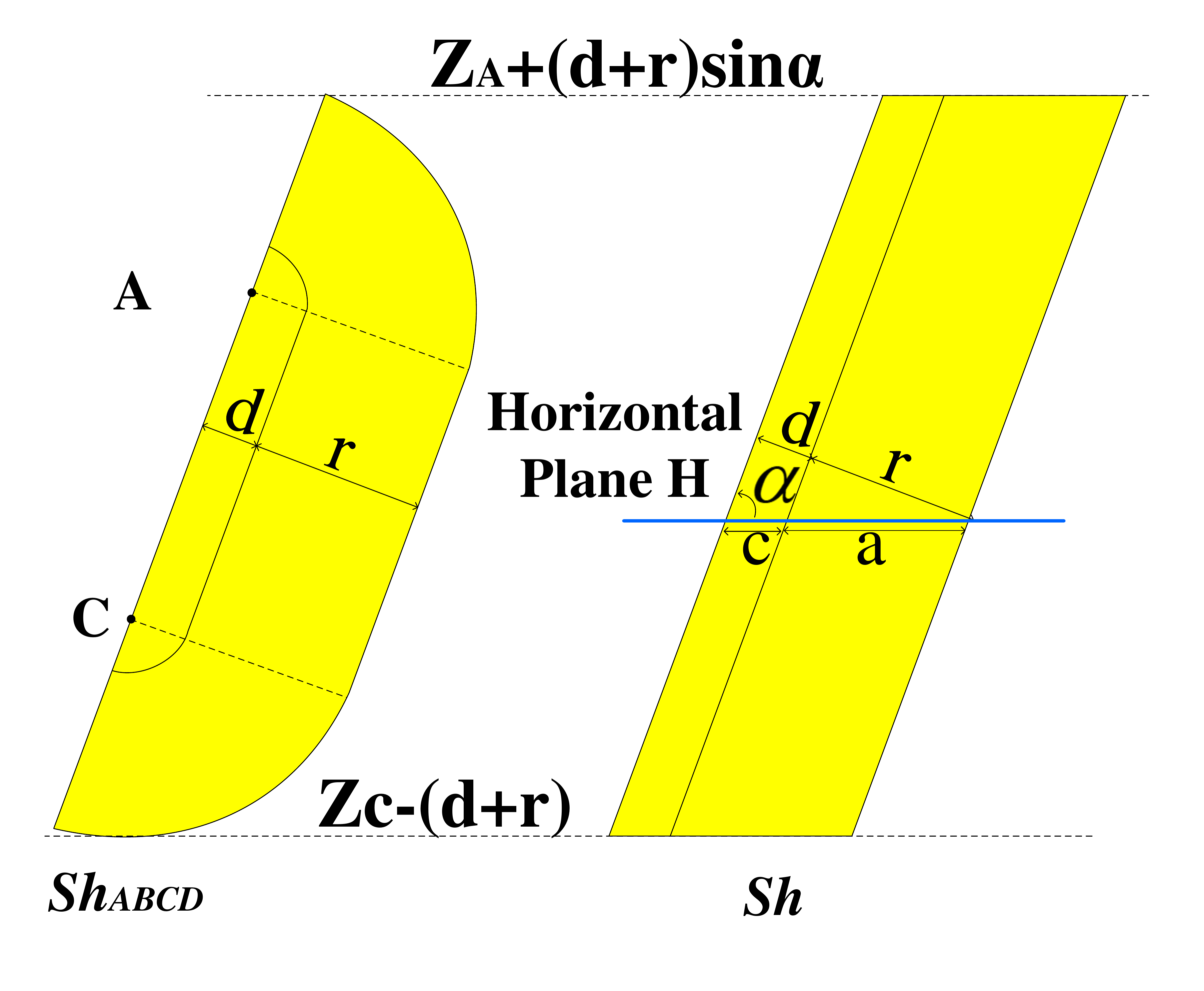}
    \label{HExtern1}
    }
  \subfigure[]{
    \includegraphics[width=1.5in]{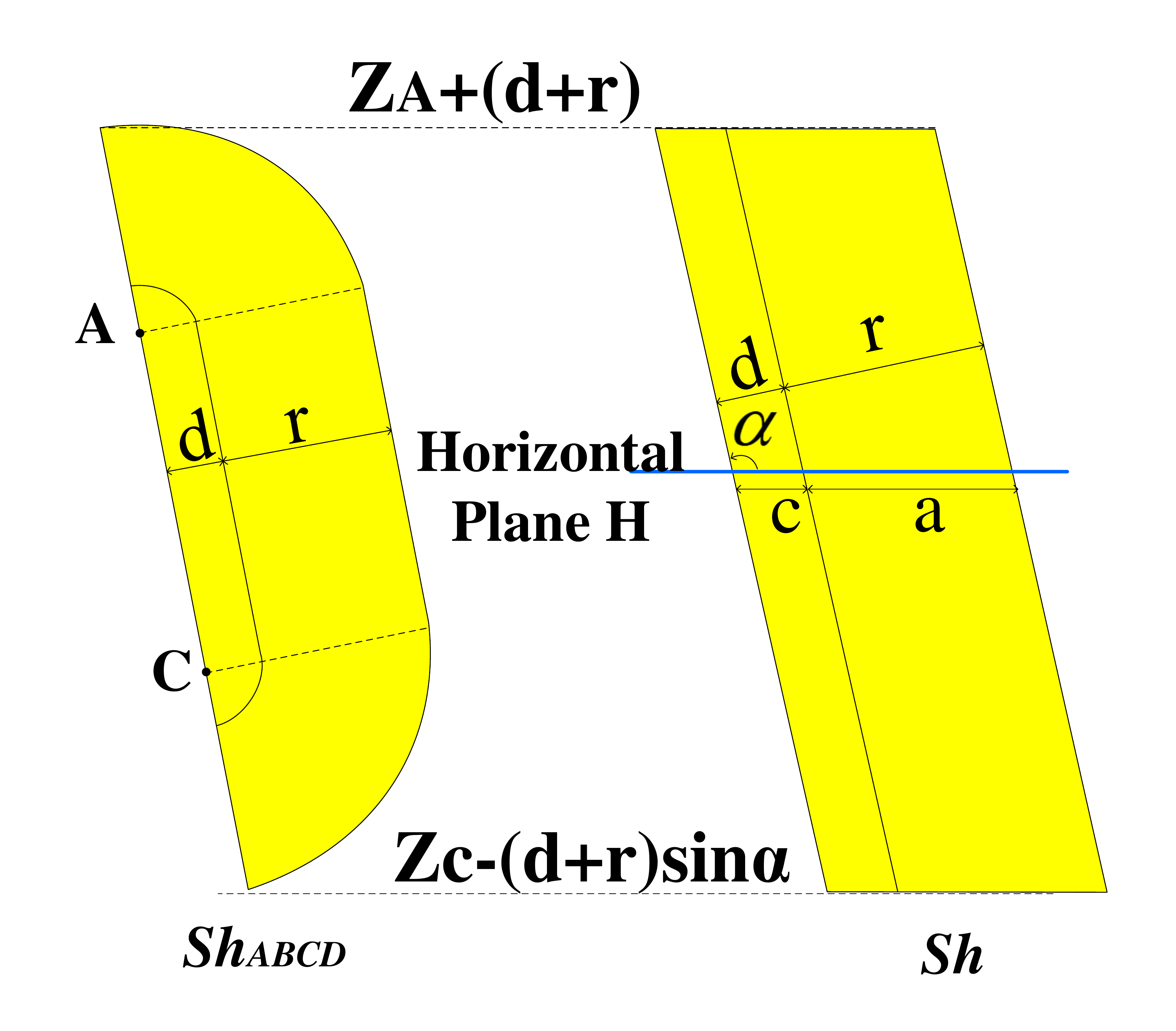}
    \label{HExtern2}
    }
    \caption{The side view of extension diagram (two cases): (a) The extension of $Sh_{ABCD}$ when $0<\alpha <\pi/2$; (b) The extension of $Sh_{ABCD}$ when $\pi/2<\alpha <\pi$.}
\end{figure}
A horizontal plane where the optical center locates is denoted as $Plane$ $H$. The section region of $Plane$ $H$ and $Sh_{ABCD}$ is denoted as $RNH$, where the camera is incapable to observe complete markers in the horizontal direction.
In order to simplify the calculation of $RNH$, the $Sh_{ABCD}$ is expanded to $Sh$ as shown in Fig. \ref{HExtern1} and \ref{HExtern2}. The space of $Sh$ is slightly larger than $Sh_{ABCD}$.
In Fig. \ref{HExtern1} and Fig. \ref{HExtern2}, left pictures are the side view  of $Sh_{ABCD}$, right pictures are the side view  of $Sh$.

\begin{figure}
  \subfigure[]{
    \includegraphics[width=1in]{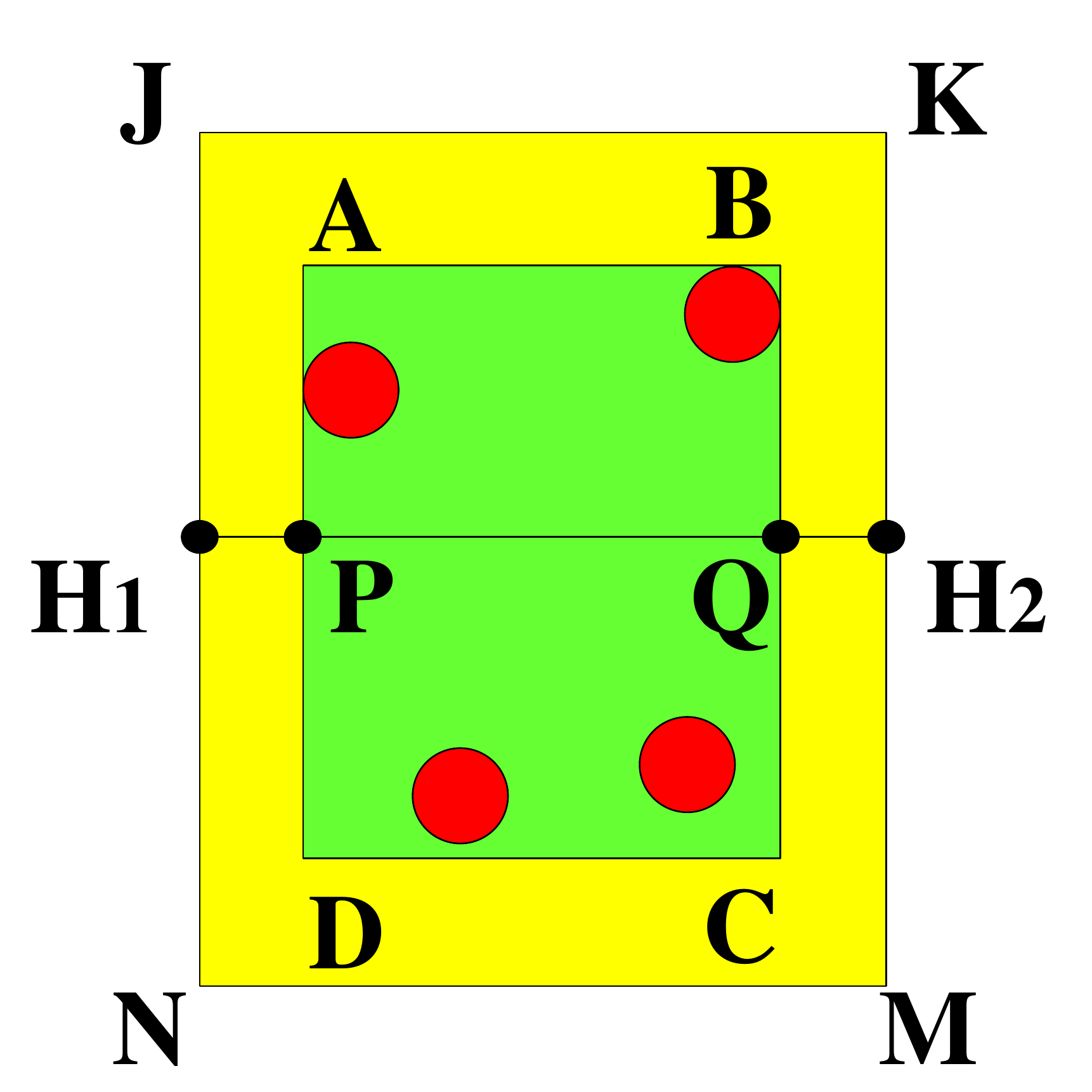}
    \label{JKMN}
    }
    \setlength{\belowdisplayskip}{3pt}
    \setlength{\belowdisplayskip}{3pt}
    \setlength{\belowdisplayskip}{3pt}
    \setlength{\belowdisplayskip}{3pt}
    \setlength{\belowdisplayskip}{3pt}
    \setlength{\belowdisplayskip}{3pt}
    \setlength{\belowdisplayskip}{3pt}
    \setlength{\belowdisplayskip}{3pt}
    \setlength{\belowdisplayskip}{3pt}
    \setlength{\belowdisplayskip}{3pt}
    \setlength{\belowdisplayskip}{3pt}
    \setlength{\belowdisplayskip}{3pt}
    \setlength{\belowdisplayskip}{3pt}
  \subfigure[]{
    \includegraphics[width=1in]{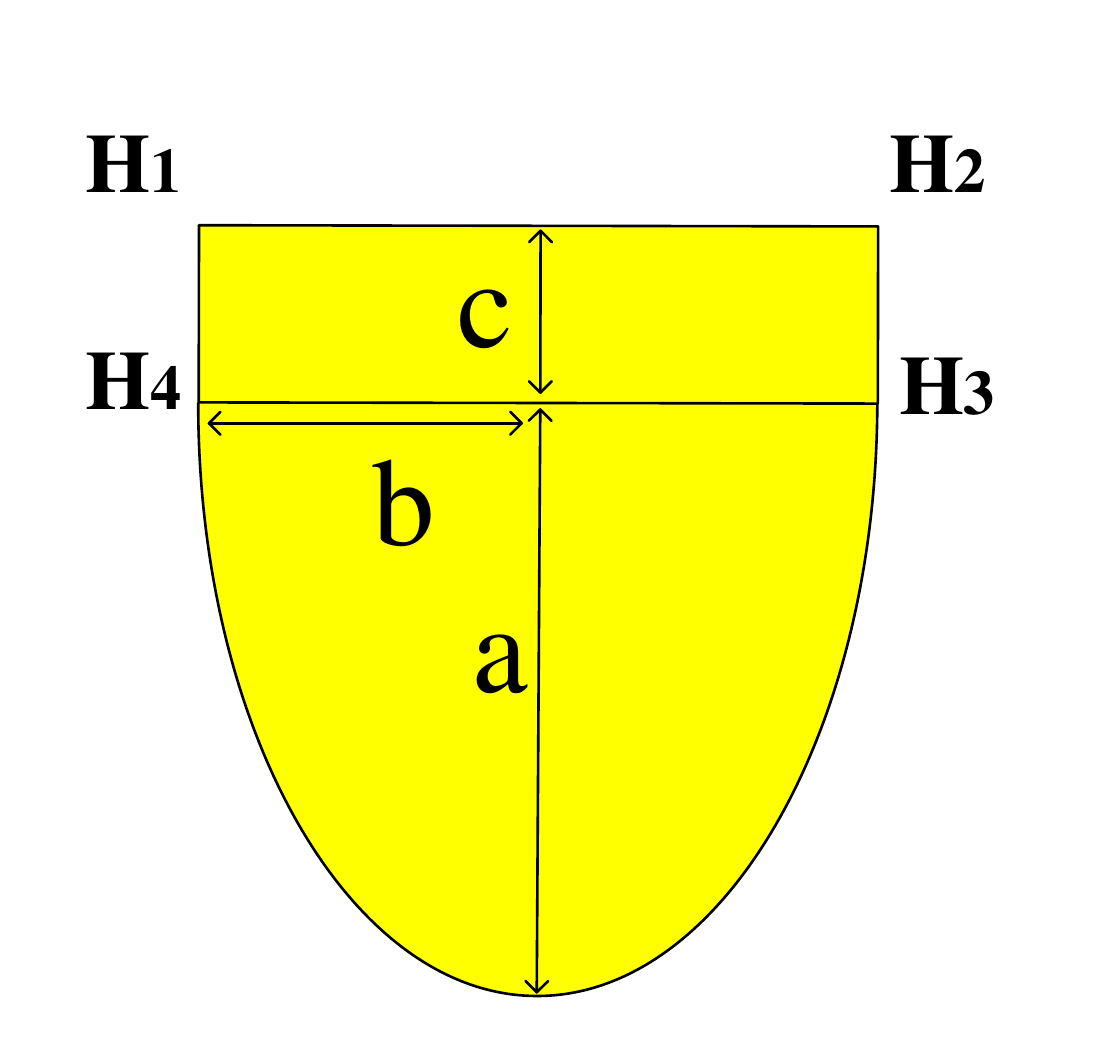}
    \label{Hqie}
    }
    \caption{(a) The front view of extension of $Sh_{ABCD}$; (b) $RNH$.}
\end{figure}

After the extension, $RNH$ can be regarded as a composition of a rectangular and a half ellipse as Fig. \ref{Hqie} shown. The length and width of the rectangle are $2b$ and $c$ respectively, and the half long axis and half short axis of the ellipse are $a$ and $b$ respectively.

In Fig. \ref{HExtern1} and \ref{HExtern2}, $\alpha$ is the inclination of the target relative to the horizontal plane. The parameter of $a$, $b$ and $c$ are obtained by equation (\ref{17})-(\ref{19}).

\begin{equation}
a=\frac{r}{\sin \alpha} \label{17}
\end{equation}
\begin{equation}
b = r \label{18}
\end{equation}
\begin{equation}
c=\frac{d}{\sin \alpha } \label{19}
\end{equation}

In Fig. \ref{JKMN}, $\overline{H_1H_2}$ is the intersection line where $Plane$ $H$ intersect with $Sh$. The section is shown in Fig. \ref{Hqie}. The coordinates of Point $H_1$ and $H_4$ can be obtained from (\ref{20})-(\ref{221}).
\begin{equation}
{H_1} = P - \frac{{2r - l}}{2}\frac{{\overrightarrow {AB} }}{{|\overrightarrow {AB} |}}\label{20}
\end{equation}
\begin{equation}
{H_2} = Q + \frac{{2r - l}}{2}\frac{{\overrightarrow {AB} }}{{|\overrightarrow {AB} |}}\label{21}
\end{equation}
where $P$ and $Q$ are the intersections of $Plane$ $H$ and $BH$.
On $Plane$ $H$, the unit vector perpendicular to $\overline{H_1H_2}$ and pointing to camera is denoted as $\overrightarrow{t_1}$.
\begin{equation}
{H_3} = {H_2} + c \cdot \overrightarrow {{t_1}}\label{22}
\end{equation}
\begin{equation}
{H_4} = {H_1} + c \cdot \overrightarrow {{t_1}}\label{221}
\end{equation}

Based on Point ${H_1}, {H_2}, {H_3}, {H_4}$ and a, b, c, the $RNH$ can be denoted.

\subsection{Vertical $FOV$ Constrain}
\begin{figure}[htbp]
    \small
    \centering
    \includegraphics[width=7cm]{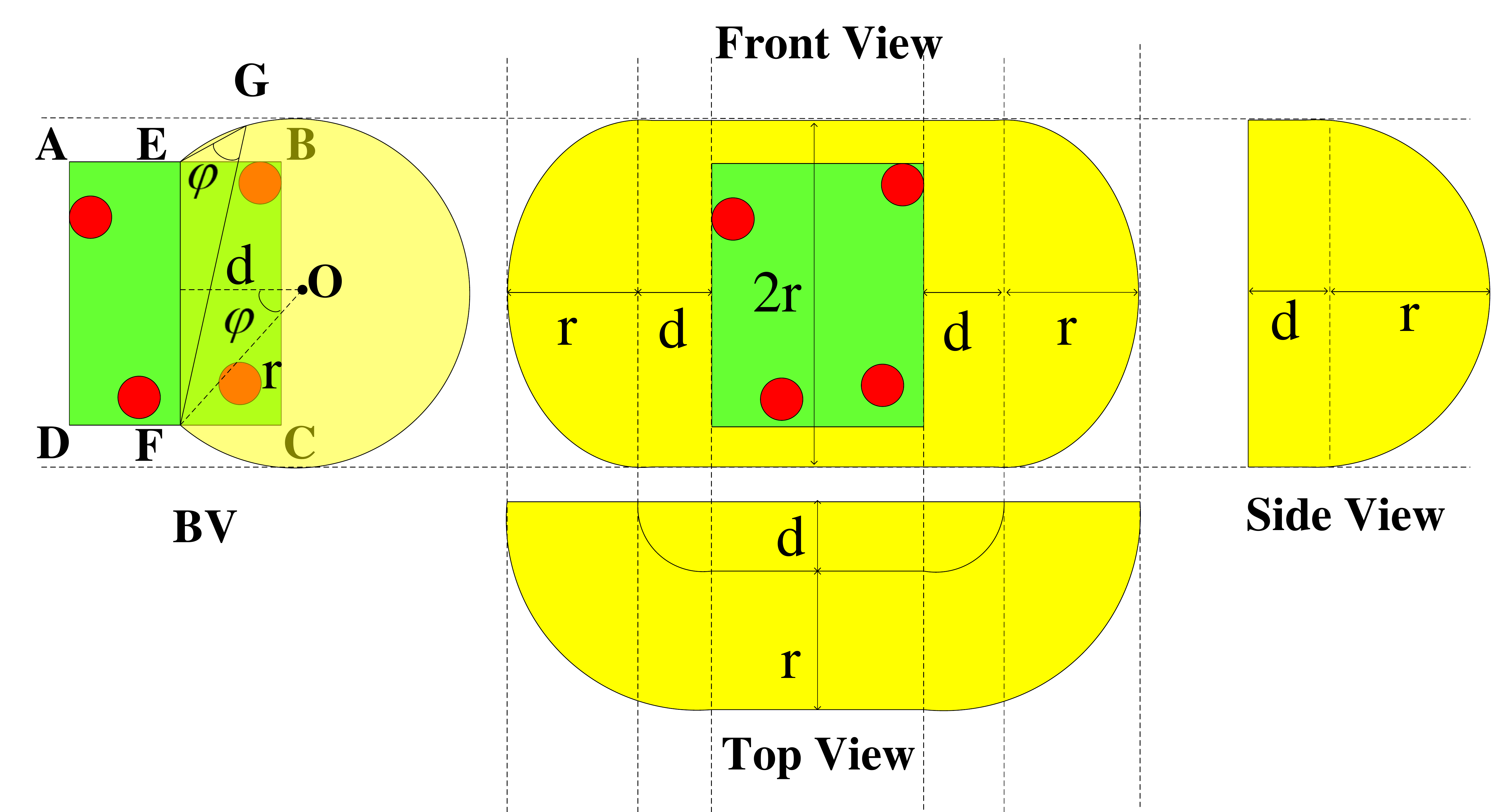}\\
    \caption{The right part of the figure is the schematic to calculate $Sv_{ABCD}$. The right part are three views of the $Sv_{ABCD}$.}
    \label{verticalConstrains}
\end{figure}

Based on $BV$, the 3D space with vertical $FOV$ constraint can be obtained. Inside this space the camera cannot keep all feature points in a view in the vertical direction. In Fig. \ref{verticalConstrains}, $Plane$ $EOF$ is perpendicular to $BV$. $\overline{EF}$ is a line segment parallel to $\overline{AD}$ and it intersects lines $\overline{AB}$  and $\overline{DC}$ at point $E$ and point $F$, respectively. For a camera (denoted as $G$) on $Plane$ $EOF$, if point $E$ and point $F$ are just kept in its view in the vertical direction, its trajectory is part of a circle, which can be proven based on $The$ $Center$ $Angle$ $Theorem$. $Plane$ $EFG$ is perpendicular to $Plane$ $ABCD$. The circle $EFG$ (the yellow region) rotates $\pm \pi/2$ around axis $\overline{EF}$, then all the covered space($Sv_{EF}$) can be obtained. Outside of $Sv_{EF}$ the camera can keep point $E$ and point $F$ in its view. Along segment $\overline{AB}$, there are infinite $\overline{EF}$ and the corresponding $Sv_{EF}$, the union of all $Sv_{EF}$ is $Sv_{ABCD}$. Outside of $Sv_{ABCD}$ the camera can keep $\overline{AB}$ and $\overline{DC}$ in its view in the vertical direction, otherwise it cannot keep them in its view in the vertical direction.

A horizontal plane where the optical center locates is denoted as $Plane$ $H$. The section of $Plane$ $H$ and $Sv_{ABCD}$ is denoted as $RNV$, where the camera is incapable to observe complete markers in vertical direction. If $Plane$ $H$ does not intersect $Sv_{ABCD}$, the $RNV$ does not exist.

In Fig. \ref{verticalConstrains}, the $r$ and $d$ of $BV$ can be obtained by Equ(\ref{23})-(\ref{25}).
\vspace{-3mm}
\begin{equation}
{l_{BC}} = \sqrt {{{({x_{\rm{B}}}{\rm{ - }}{x_{\rm{C}}})}^2}{\rm{ + }}{{({{\rm{y}}_{\rm{B}}}{\rm{ - }}{y_{\rm{C}}})}^2}{\rm{ + }}{{({z_{\rm{B}}}{\rm{ - }}{z_{\rm{C}}})}^2}} \label{23}
\end{equation}
\begin{equation}
r = \frac{{{l_{BC}}}}{{2\sin \theta }}\label{24}
\end{equation}
\begin{equation}
d = \frac{{{l_{BC}}}}{{2\tan \theta }}\label{25}
\end{equation}

\begin{figure}[h]
 \subfigure[]{
    \includegraphics[width=1.1in]{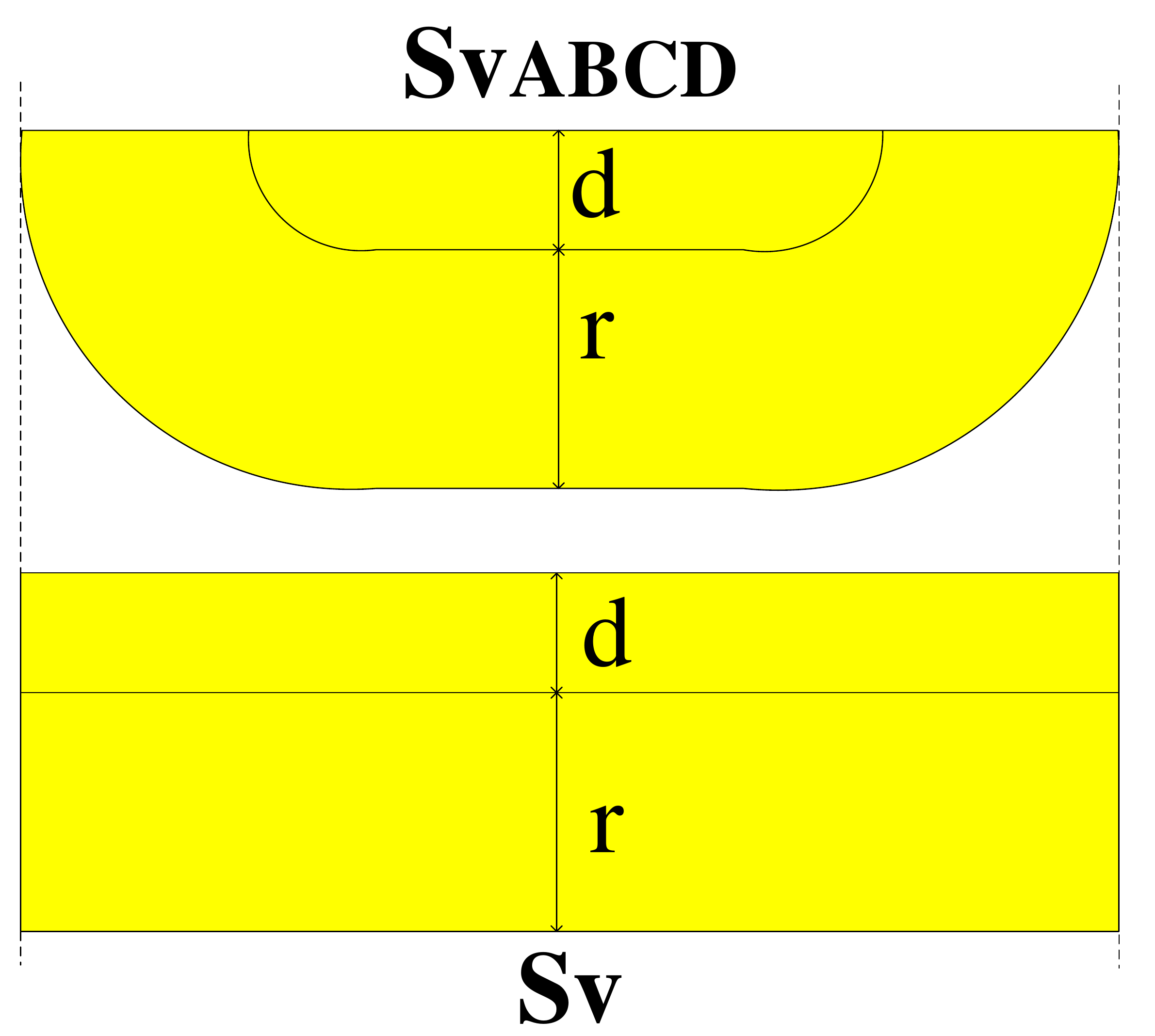}
    \label{verticalSectionSimple}
    }
  \subfigure[]{
    \includegraphics[width=1.2in]{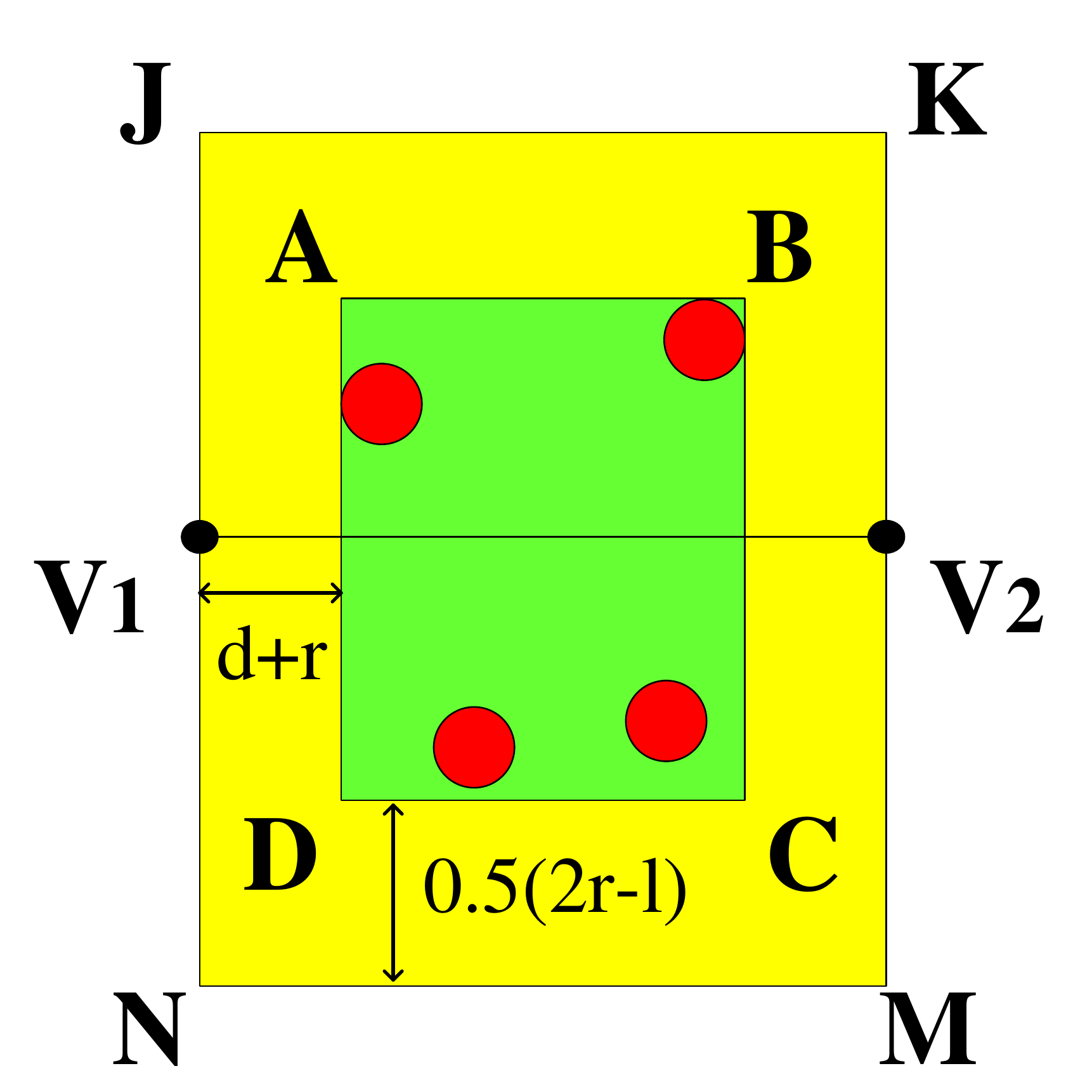}
    \label{VJKMN}
    }
  \subfigure[]{
    \includegraphics[width=1in]{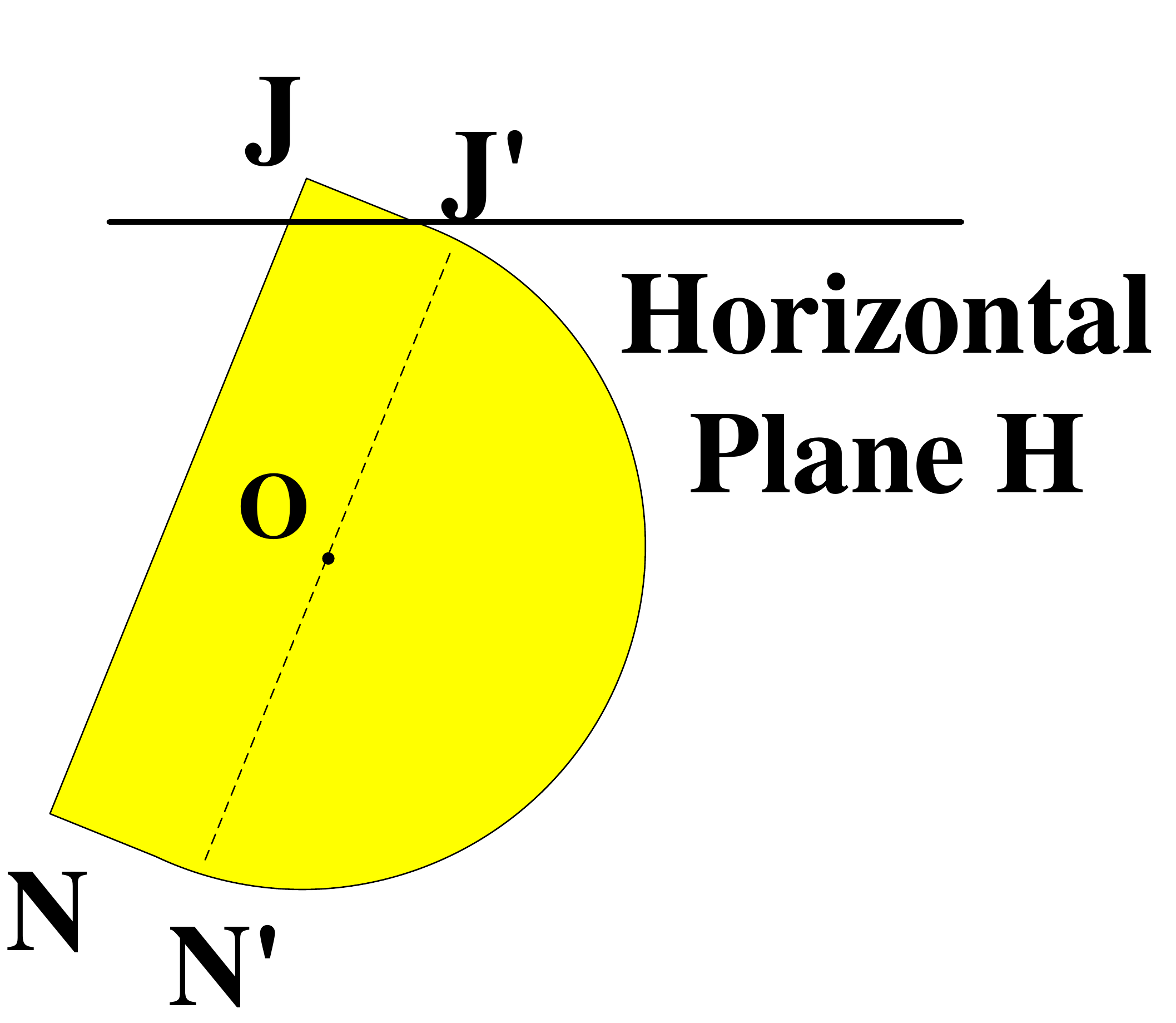}
    \label{Vqie1}
    }
    \setlength{\belowdisplayskip}{3pt}
    \setlength{\belowdisplayskip}{3pt}
    \setlength{\belowdisplayskip}{3pt}
    \setlength{\belowdisplayskip}{3pt}
    \setlength{\belowdisplayskip}{3pt}
    \setlength{\belowdisplayskip}{3pt}
    \setlength{\belowdisplayskip}{3pt}
    \setlength{\belowdisplayskip}{3pt}
    \setlength{\belowdisplayskip}{3pt}
    \setlength{\belowdisplayskip}{3pt}
    \setlength{\belowdisplayskip}{3pt}
    \setlength{\belowdisplayskip}{3pt}
    \setlength{\belowdisplayskip}{3pt}
    \setlength{\belowdisplayskip}{3pt}
    \setlength{\belowdisplayskip}{3pt}
    \setlength{\belowdisplayskip}{3pt}
    \setlength{\belowdisplayskip}{3pt}
    \setlength{\belowdisplayskip}{3pt}
    \setlength{\belowdisplayskip}{3pt}
    \setlength{\belowdisplayskip}{3pt}
    \setlength{\belowdisplayskip}{3pt}
    \setlength{\belowdisplayskip}{3pt}
    \setlength{\belowdisplayskip}{3pt}
    \subfigure[]{
    \includegraphics[width=1in]{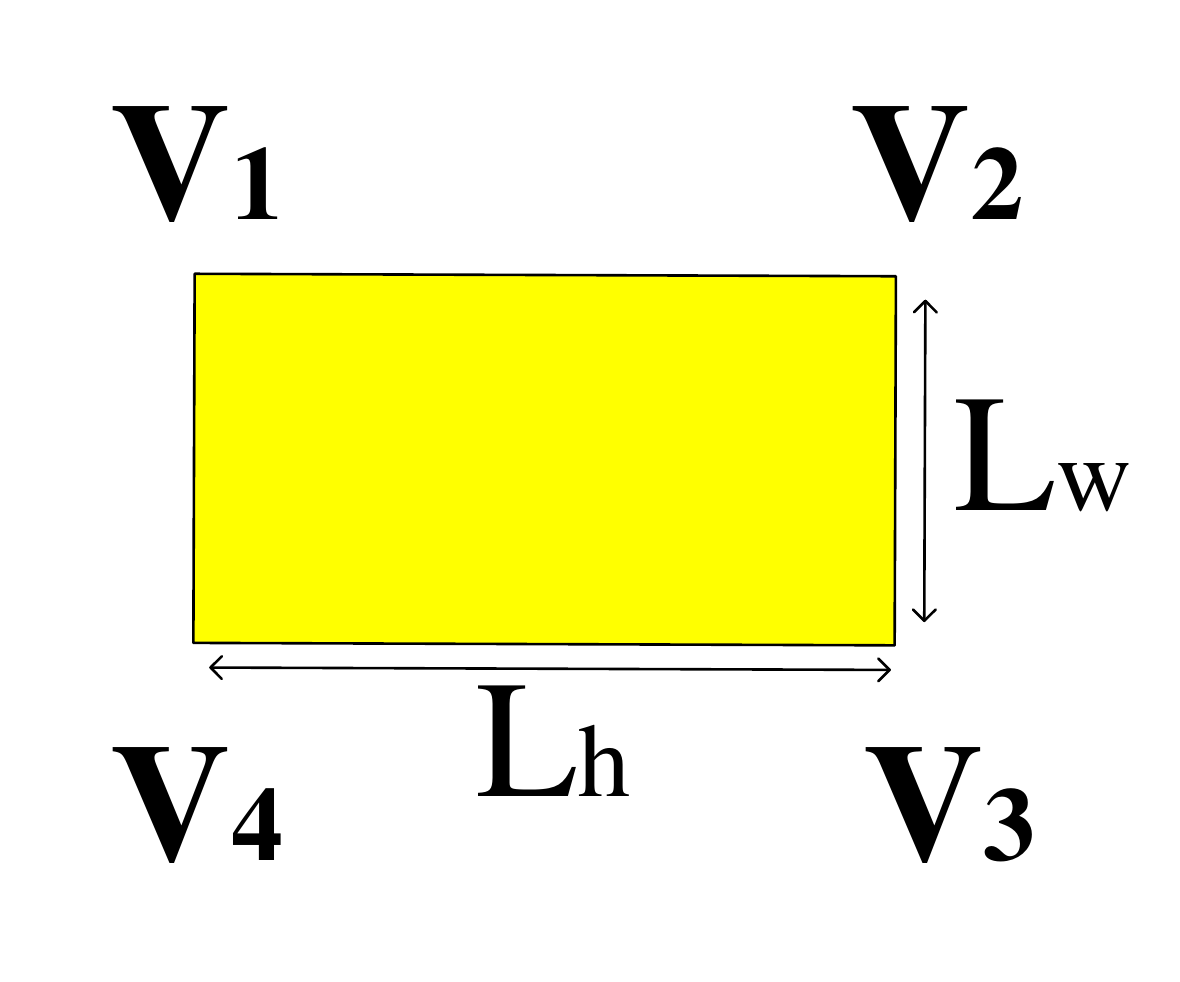}
    \label{Vqie}
    }
    \caption{(a) The top part of the figure is the top view of $Sv_{ABCD}$, and the bottom part of the figure is the top view of $Sv$. (b) The front view of $Sv$.
    (c) The side view of schematic to calculate $RNV$ based on $Sv$ when ${J_z}^\prime  \le h_c < {J_z}$.
    (d) $RNV$.}
\end{figure}

In order to simplify calculation, the $Sv_{ABCD}$ is expended to $S_v$ as Fig. \ref{verticalSectionSimple} shown, the top picture is $Sv_{ABCD}$, and the bottom picture is $S_v$. If $Plane H$ intersects $Sv$, the $RNV$ can be obtained. The front view of $Sv$ is shown in Fig. \ref{VJKMN}, $Plane$ $ABCD$ and $Plane$ $JKMN$ are coplanar. Some parameters used to describe $S_v$ can be calculated by Equ(\ref{261})-(\ref{311}).

\vspace{-7mm}
\vspace{-5mm}
\begin{equation}
J = A - (d + r)\cdot\frac{{\overrightarrow {AB} }}{{|\overrightarrow {AB} |}} - \frac{{2r - l}}{2}\cdot\frac{{\overrightarrow {AD} }}{{|\overrightarrow {AD} |}}\label{261}
\end{equation}
\begin{equation}
K = B + (d + r)\cdot\frac{{\overrightarrow {AB} }}{{|\overrightarrow {AB} |}} - \frac{{2r - l}}{2}\cdot\frac{{\overrightarrow {AD} }}{{|\overrightarrow {AD} |}}\label{281}
\end{equation}
\begin{equation}
N = D - (d + r)\cdot\frac{{\overrightarrow {AB} }}{{|\overrightarrow {AB} |}} + \frac{{2r - l}}{2}\cdot\frac{{\overrightarrow {AD} }}{{|\overrightarrow {AD} |}}\label{301}
\end{equation}
\begin{equation}
M = C + (d + r)\cdot\frac{{\overrightarrow {AB} }}{{|\overrightarrow {AB} |}} + \frac{{2r - l}}{2}\cdot\frac{{\overrightarrow {AD} }}{{|\overrightarrow {AD} |}}\label{321}
\end{equation}


\begin{equation}
{J^\prime } = J + d\cdot \overrightarrow {{bv_1}}, \qquad {K^\prime } = K + d\cdot \overrightarrow {{bv_1}}
\label{271}
\end{equation}
\begin{equation}
{N^\prime } = N + d\cdot \overrightarrow {{bv_1}}, \qquad  {M^\prime } = M + d\cdot \overrightarrow {{bv_1}}\label{311}
\end{equation}
where $J^\prime$, $K^\prime$, $N^\prime$ and $M^\prime$ are points of $J$, $K$, $N$ and $M$ in the direction of $\overrightarrow {{bv_1}}$.
Now the coordinates of $J$, ${J^\prime}$, ... ,$M$, ${M^\prime }$ can be obtained, and the values of $a$, $b$, $r$ are known. $h_c$ denotes the $z$ coordinate of $Plane$ $H$ in the mobile robot coordinate system. According to $h_c$, the section $RNV$ of $S_v$ and $Plane$ $H$ can be obtained, and its edge of $RNV$ can be obtained. The edge of $RNV$ is denoted by $V_{1-4}$. When the inclination $\alpha$ is acute, there are five cases.

The first case is shown in Fig. \ref{Vqie1}, it is the case when ${J_z}^\prime  \le h_c < {J_z}$ where ${J_z}$ and ${J_z}^\prime$ are the z-coordinates of points $J$ and ${J}^\prime$ respectively. $V_1$ and $V_4$ are the intersection points between $Plane H$ and lines $JN$, $JJ^\prime$, respectively. $V_2$ and $V_3$ are the intersection points of $Plane H$ and lines $KM$, $KK^\prime$, respectively. The vector from point $N$ to $J$ is denoted as $\overrightarrow p$, and its z-coordinate is represented as $p_z$. The $V_1$ can be obtained by equation (\ref{26})-(\ref{28}). And it is similar to solve $V_{2-4}$. Other cases can be obtained based on some basic geometric relations.
\begin{equation}
\overrightarrow p  = N - J \label{26}
\end{equation}
\begin{equation}
h_c = {J_z} + m_3\cdot  {{p_z}}\label{27}
\end{equation}
\begin{equation}
{V_1} = J + m_3\cdot \overrightarrow p\label{28}
\end{equation}
where $m_3$ is a coefficient to be determined.
\subsection{The Union Region of $RNH$ and $RNV$.}
\begin{figure}
  \subfigure[]{
    \includegraphics[width=1.45in]{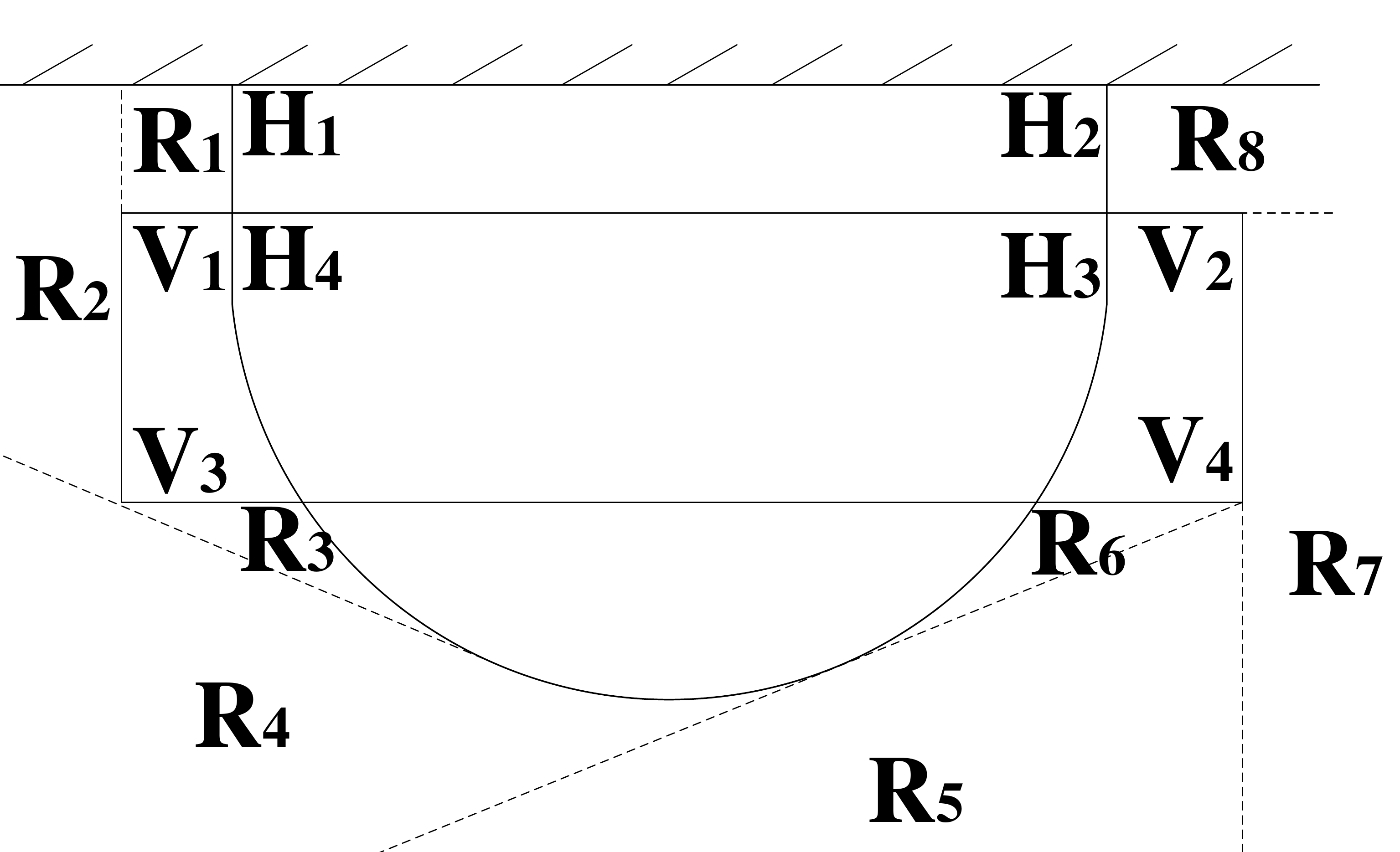}
    \label{TheShortestPaths1}
    }
    \subfigure[]{
    \includegraphics[width=1.45in]{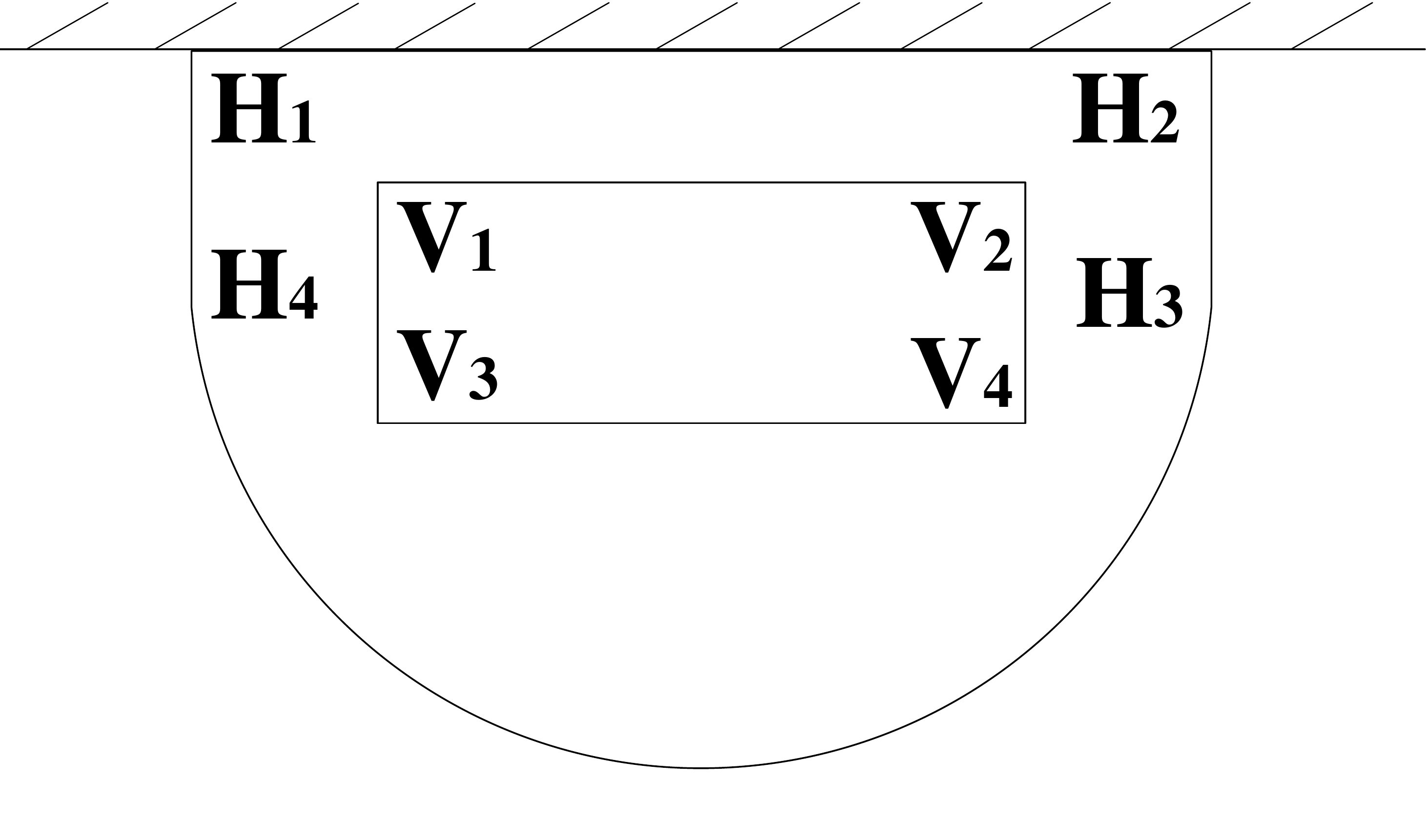}
    \label{TheShortestPaths2}
    }
    \subfigure[]{
    \includegraphics[width=1.45in]{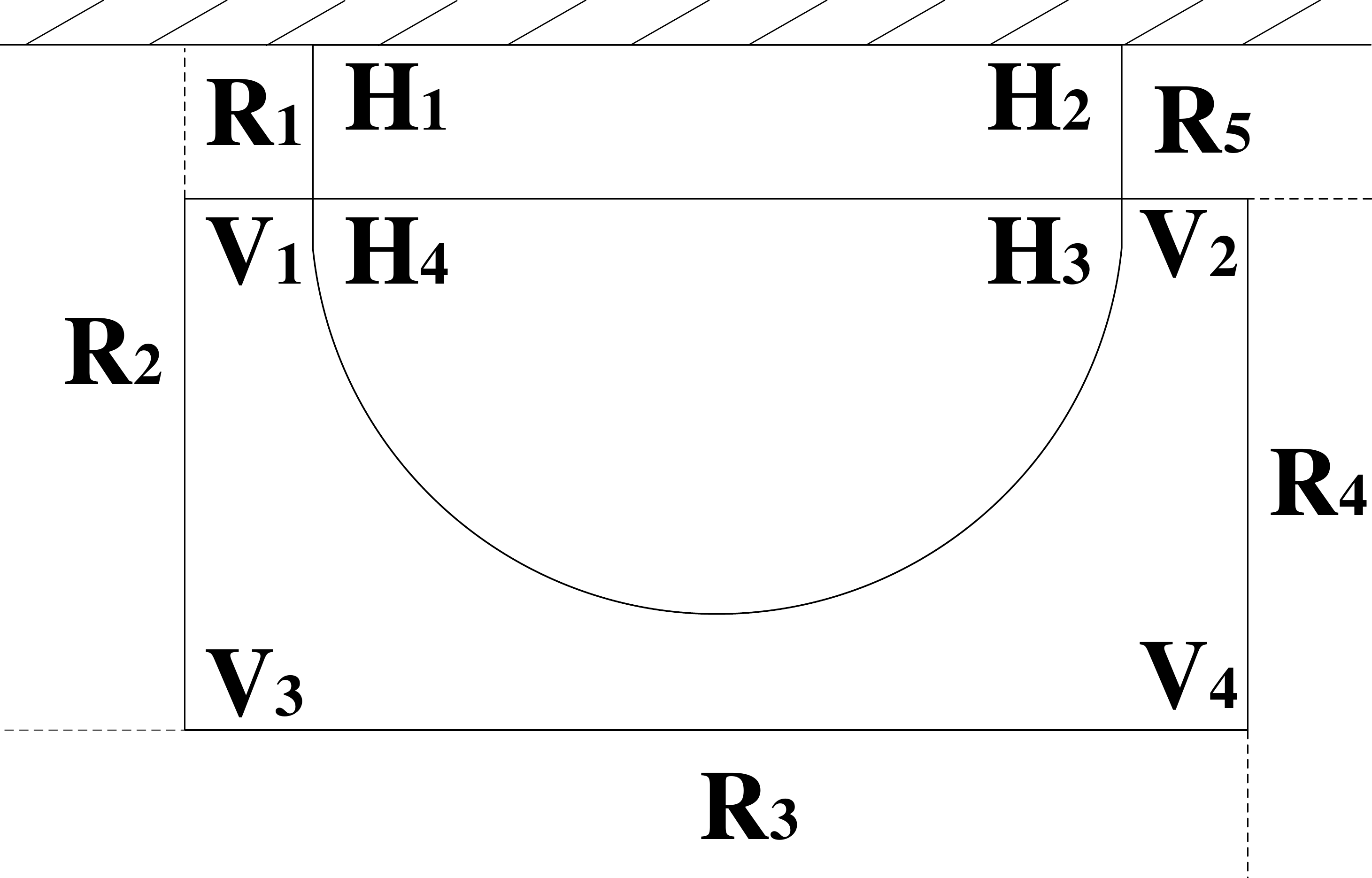}
    \label{TheShortestPaths3}
    }
    \setlength{\belowdisplayskip}{3pt}
    \setlength{\belowdisplayskip}{3pt}
    \setlength{\belowdisplayskip}{3pt}
    \setlength{\belowdisplayskip}{3pt}
    \subfigure[]{
    \includegraphics[width=1.45in]{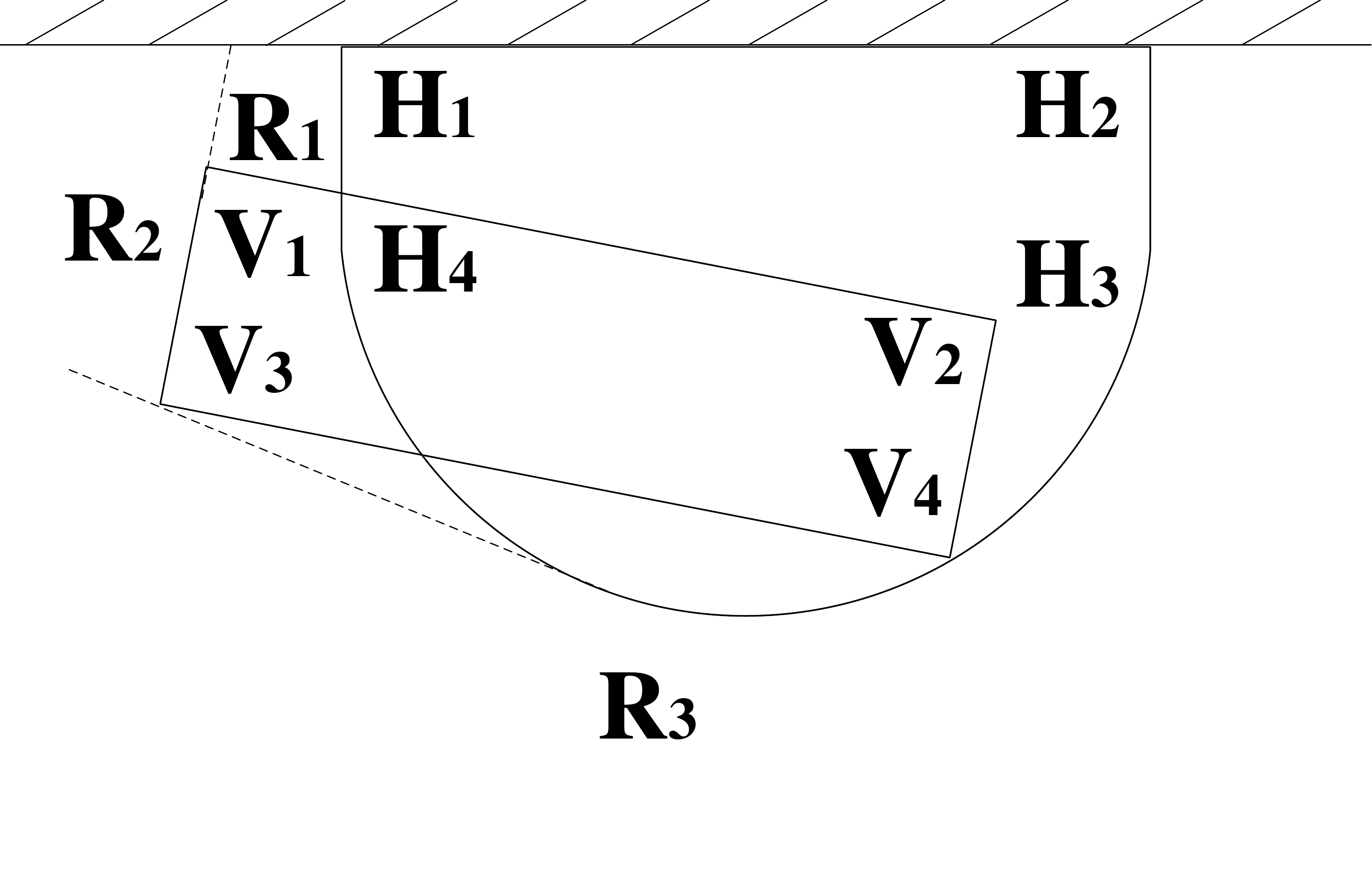}
    \label{TheShortestPaths4}
    }
    \caption{Four kinds shape of $RNA$.}
\end{figure}

After obtaining $RNH$ and $RNV$, the $RNA$ can be acquired by the union of $RNH$ and $RNV$. According to the actual situation, they are mainly divided into these four cases shown in Fig. \ref{TheShortestPaths1}-\ref{TheShortestPaths4}.

%
%

\section{Simulation And Experiment}
 In this section two simulations are conducted.
 The first simulation is used to verify the effectiveness of our approach by virtual images. When the camera locates different positions, the feature points will approach edge of image in the horizontal or vertical direction. The second simulation records the curves of distances between feature points and the edge of image at pixel level when the mobile robot moves. This simulation can prove that under the guidance of $RNA$ the camera can keep markers in a view during the motion of mobile robot.

   $MATLAB$ is used to assist the simulations, the horizontal angular aperture $\theta$ and the vertical angular aperture $\varphi$ of the virtual camera are both 1.13 rad. The image size is 1024$\times$1024.

\begin{figure}
  \subfigure[]{
    \includegraphics[width=1.45in]{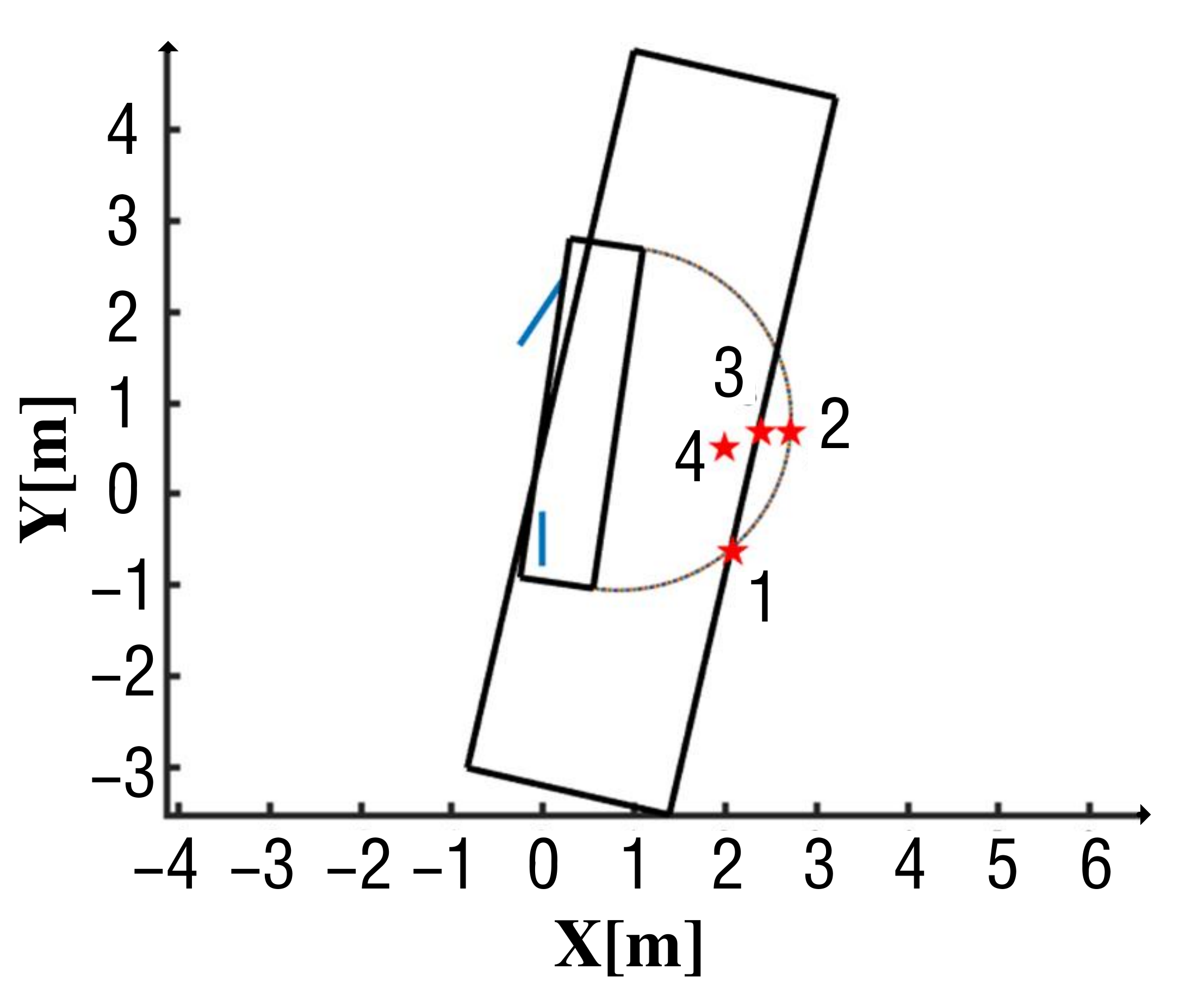}
    \label{all}
    }
        \subfigure[]{
    \includegraphics[width=1.45in]{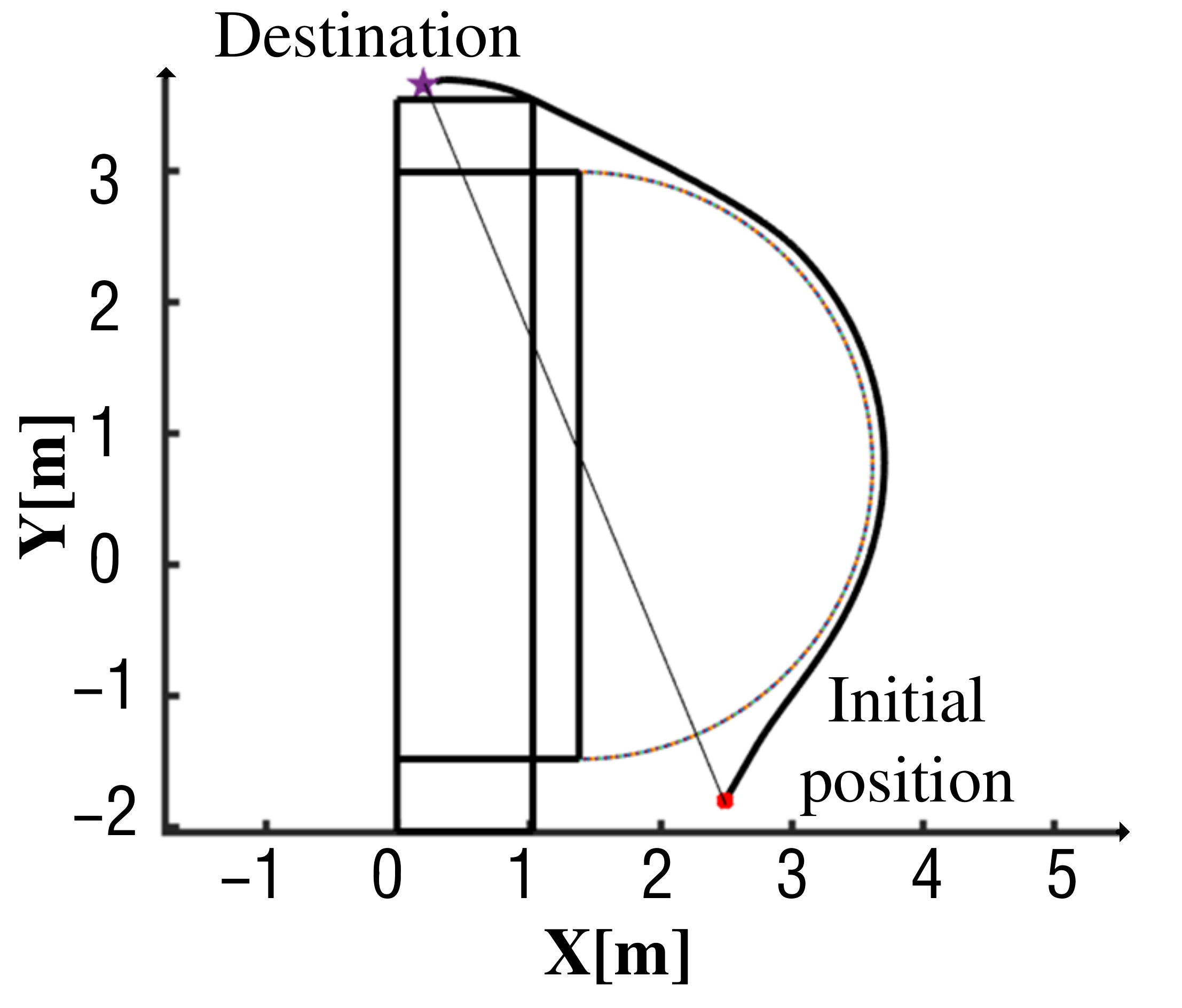}
    \label{Sim1FOVConstraint}
    }
    \subfigure[]{
    \includegraphics[width=1.5in]{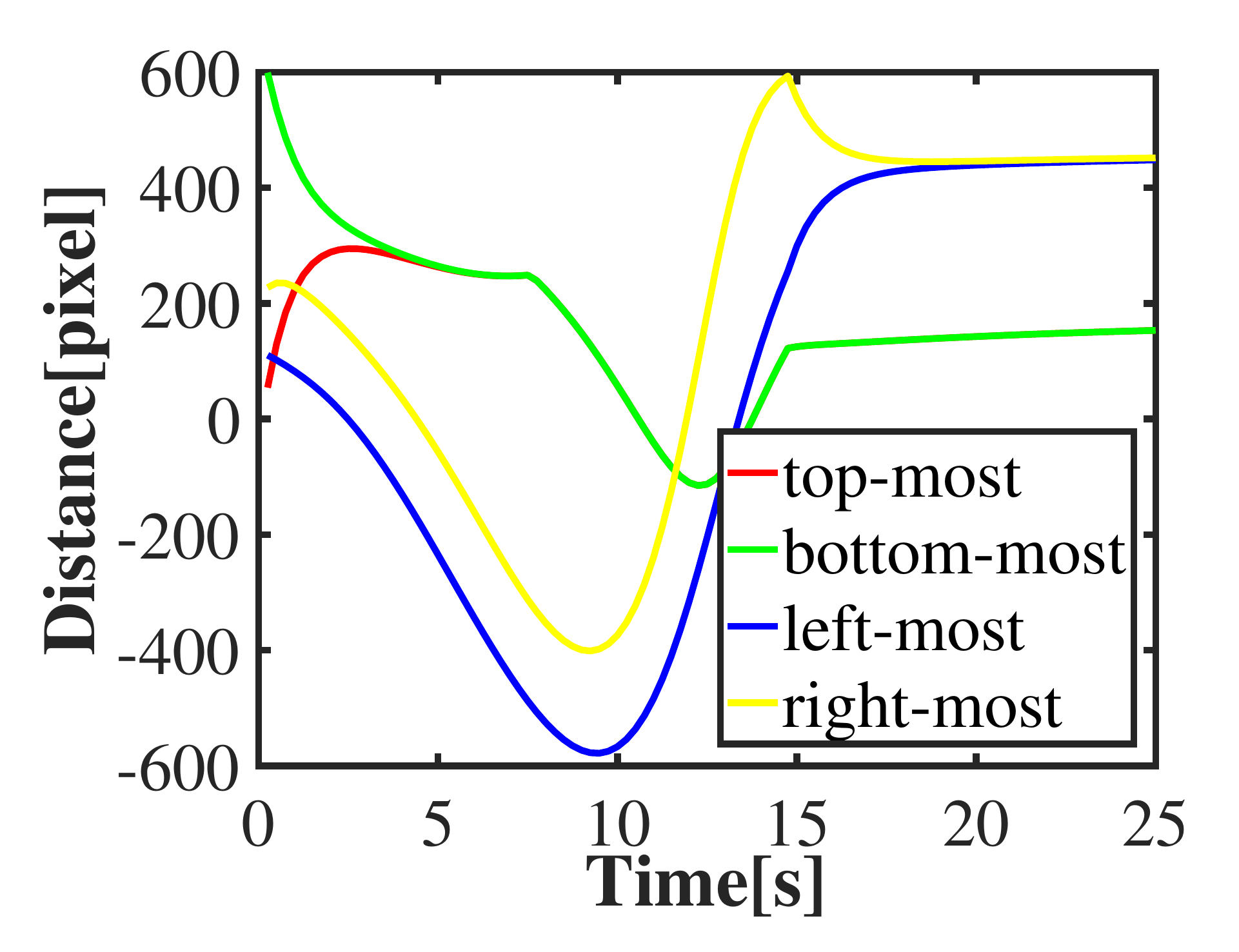}
    \label{Sim1StraightLinePixDis}
    }
    \subfigure[]{
    \includegraphics[width=1.5in]{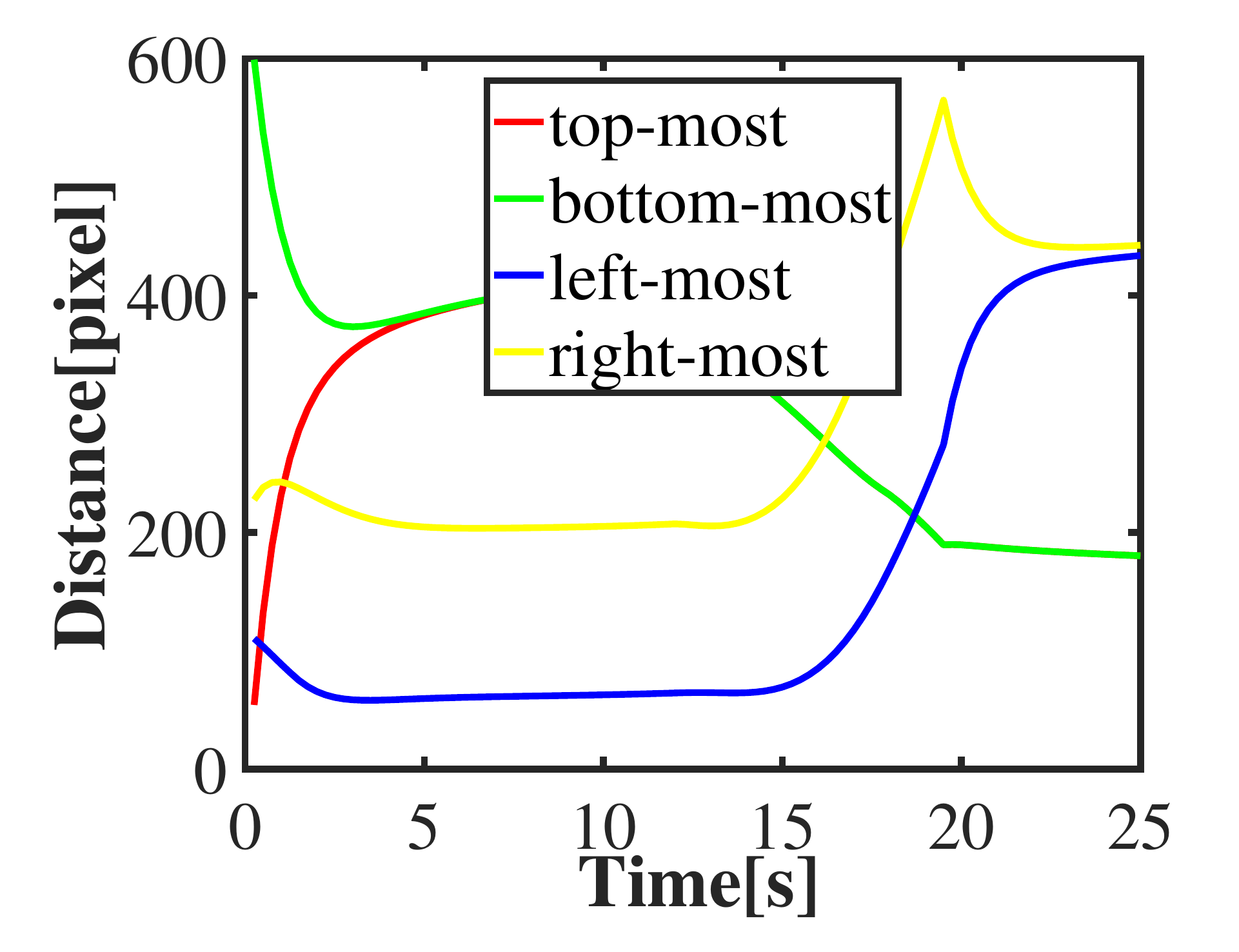}
    \label{Sim1TrajectoryPlanPixDis}
    }
  \subfigure[]{
    \includegraphics[width=1in]{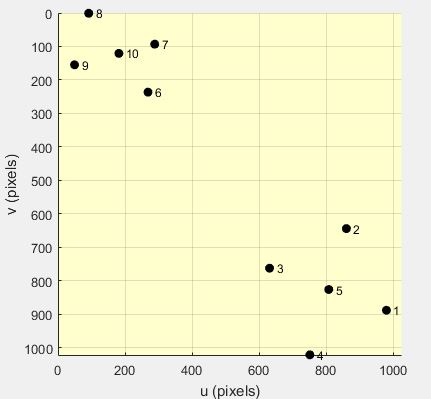}
    \label{Position1View}
    }
     \subfigure[]{
    \includegraphics[width=1in]{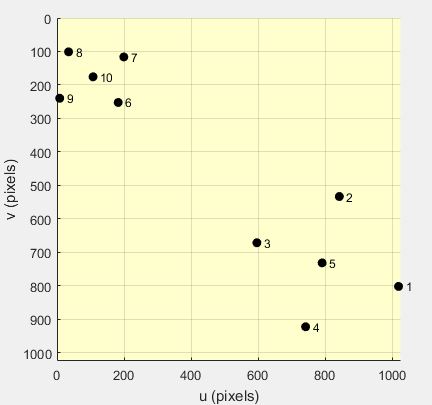}
    \label{Position2View}
    }
     \subfigure[]{
    \includegraphics[width=1in]{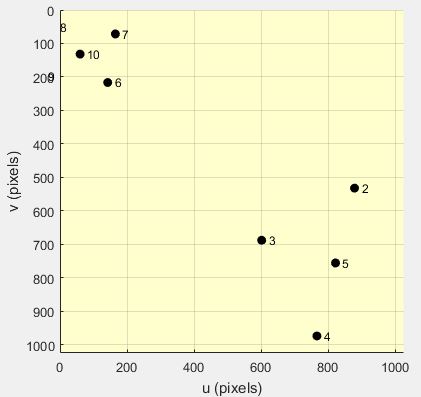}
    \label{Position3View}
    }
    \setlength{\belowdisplayskip}{3pt}
    \setlength{\belowdisplayskip}{3pt}
    \setlength{\belowdisplayskip}{3pt}
    \setlength{\belowdisplayskip}{3pt}
    \setlength{\belowdisplayskip}{3pt}
    \setlength{\belowdisplayskip}{3pt}
    \setlength{\belowdisplayskip}{3pt}
    \setlength{\belowdisplayskip}{3pt}
    \setlength{\belowdisplayskip}{3pt}
    \setlength{\belowdisplayskip}{3pt}
    \setlength{\belowdisplayskip}{3pt}
     \subfigure[]{
    \includegraphics[width=1in]{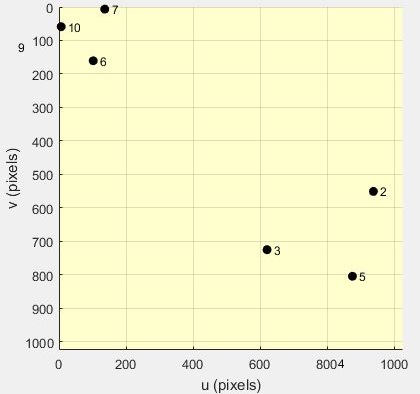}
    \label{Position4View}
    }

    \caption{(a) The shape of $RNA$ and the position 1-4 of mobile robot (pentagram). (b) The shape of $RNA$ and the trajectory of mobile robot. (c) The pixel distance when the mobile robot moves along straight line in Fig.\ref{Sim1FOVConstraint}. (d) The pixel distance when the mobile robot moves along the edge of the $RNA$ in Fig.\ref{Sim1FOVConstraint}. (e-f) The image taken at the Position 1-4 in Fig.\ref{all}.}
\end{figure}

In the first simulation, there are two markers and each marker has five points. The center coordinates of the two targets are (0,2,3) and (0,-0.5,5) in the world coordinate system, respectively. By the above method, the $RNV$ and $RNA$ can be calculated, and they are drawn in Fig.\ref{all}. The blue lines are two targets. In Fig. \ref{all}, the mobile robot locates at four different position, which are marked with number 1-4.
Position 1 is the intersection of region $RNA$ and region $RNV$, and the image captured by camera at position 1 is shown in Fig. \ref{Position1View}. From this figure it can be seen that the left-most and right-most points are close to the edge of image in the horizontal direction. The top-most and bottom-most points are close to the edge of image in the vertical direction. If the camera pan or tilt any tiny angle, the point near the edge of image will be easily out of view.
Position 2 locates the edge of $RNH$, and the image is shown in Fig.\ref{Position2View}. The image coordinates of the left-most point and the right-most are (6, 24) and (1018, 803), respectively. They are very close to the edge of the image in the horizontal direction.
Position 3 is at the edge of $RNV$. In Fig .\ref{Position3View},
the top-most point and bottom-most points are (-15, 44) and (767, 974), respectively. Because of $0<44<974<1024$, all points can be kept in a view in the vertical direction. It also can be seen that the left-most and right-most points are out of view in the horizontal direction.
Position 4 is inside $RNH$ and $RNA$. Fig .\ref{Position4View} is taken at position 4, and it is obvious that the camera cannot keep all points in a view at this position.

Based on the simulation, it can be known that when the camera locates at the edge of $RNH$, the points are close to the edge of image in the horizontal direction. When the camera locates at the edge of $RNV$, the points are close to the edge of image in the vertical direction. They are consistent with the results inferred from the $FOV$ constraint region $RNA$. They prove the effectiveness of $RNA$.

%
%
%
%

In the second simulation, the mobile robot moves along a curve and a straight paths, respectively. Their initial position are both (2.2, -1.8, 0), and destinations are both (0.2, 3.65, 0) as shown in Fig. \ref{Sim1FOVConstraint}. The curve path is around the edge of $RNA$. During the motion of the mobile robot, the pixel distances between feature points and edges of image are recorded as shown in Fig. \ref{Sim1StraightLinePixDis} and Fig. \ref{Sim1TrajectoryPlanPixDis}. The positive values of left-most and right-most mean that all points cannot leave the field of vision in the horizontal direction, and the positive values of top-most and bottom-most mean that all points cannot leave the field of vision in the vertical direction.
%
Along the straight path in Fig. \ref{Sim1FOVConstraint}, the mobile robot goes into the $RNH$ and then goes into $RNV$, finally it arrives at the destination. At 2.25s, the mobile robot arrives at (2.452, -1.115, 0) which is the edge of $RNH$. From Fig.\ref{Sim1StraightLinePixDis}, it can be seen the pixel distance of right-most and left-most are 16 and 164 respectively, the pixel distance of top-most and bottom-most are 293 and 342 respectively. It can be seen that the right-most and left-most points are close to the edge of images. At 9.25s the values of right-most and left-most points are -401 and -578, respectively. It means that the points are out of view in the horizontal direction.
 Then the mobile robot arrives at the edge of $RNV$ at 10.5s, and the values of top-most and bottom-most all become 8. It can be known that the top-most and bottom-most points are close to the edges of images. At 12.25s the values of top-most and bottom-most are both -115. This illustrates that points are out of view in the vertical direction. The destination is outside of the $RNH$ and $RNV$. Finally the mobile robot reaches the destination, and all points can be kept in the camera's view again.
Fig. \ref{Sim1TrajectoryPlanPixDis} shows the changes of pixel distance when the mobile robot moves along the edge of $RNA$. In this figure all values are positive and the minimum pixel distance is 59, so all points are always kept in the camera's view.

From the pixel distance of the straight line, one can see that when the mobile robot moves into the $RNH$, feature points are out of view in the horizontal direction, when the mobile robot moves into the $RNV$, feature points are out of view in the vertical direction. It further verifies the effectiveness of our method. Under the guidance of $RNA$, the mobile robot with pan-tilt camera can arrive at the destination with a curve path, which can keep all feature point in a view. It means that $RNA$ can guide the motion of mobile robot effectively.
\begin{figure}
\subfigure[]{
\includegraphics[width=1.45in]{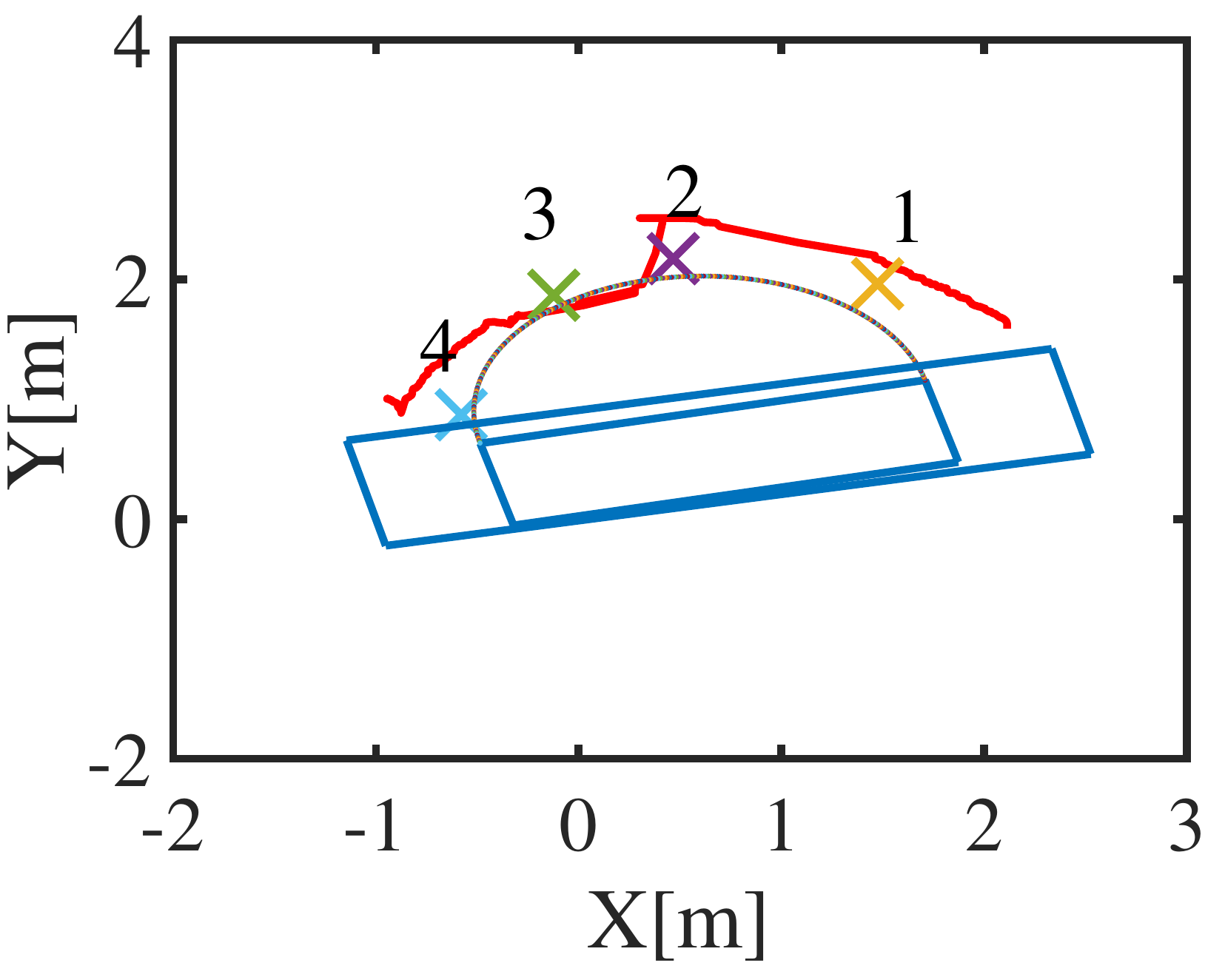}
\label{TestTrajectory}
}
\subfigure[]{
\includegraphics[width=1.45in]{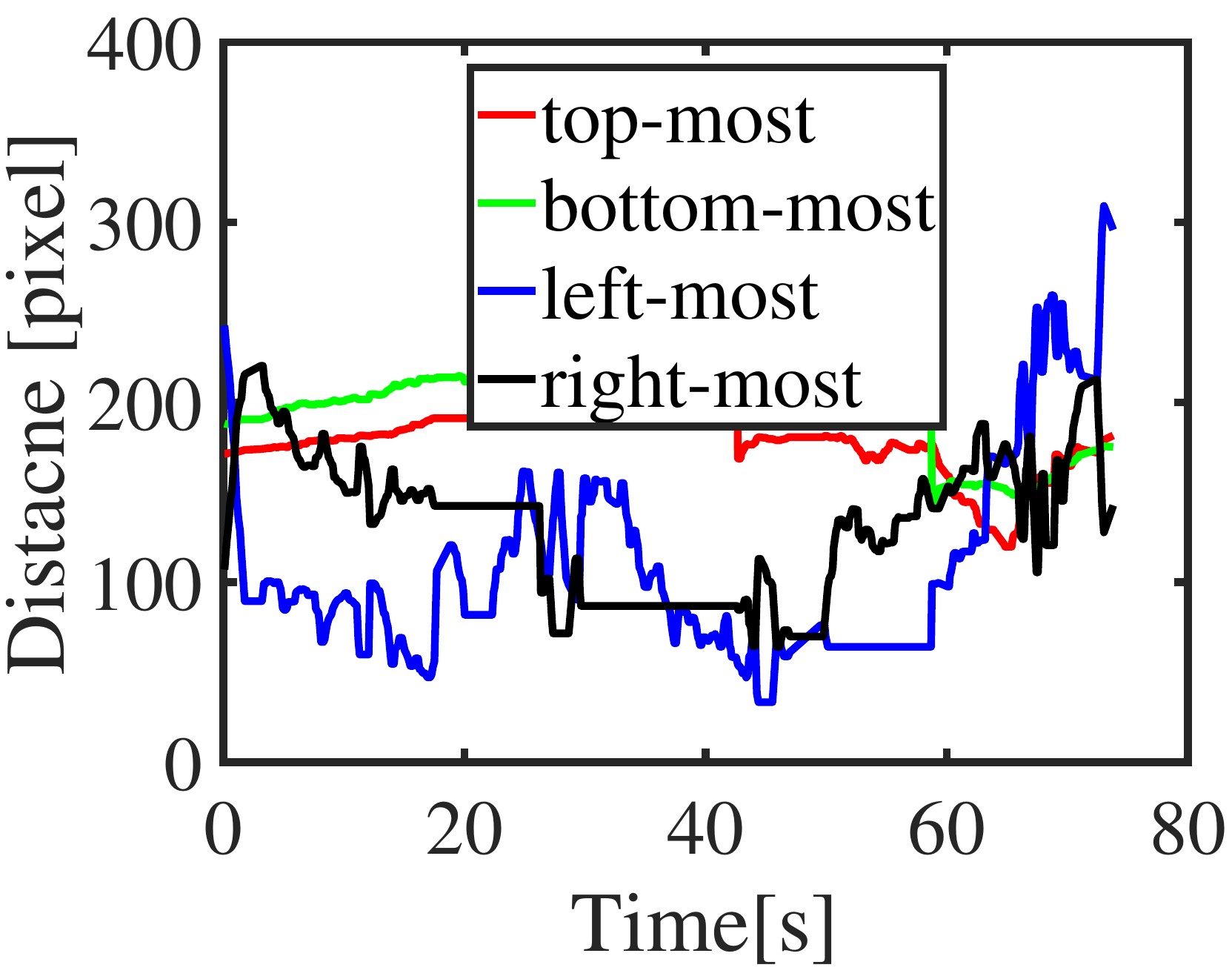}
\label{TestPixdis}
}
\subfigure[]{
\includegraphics[width=0.9in]{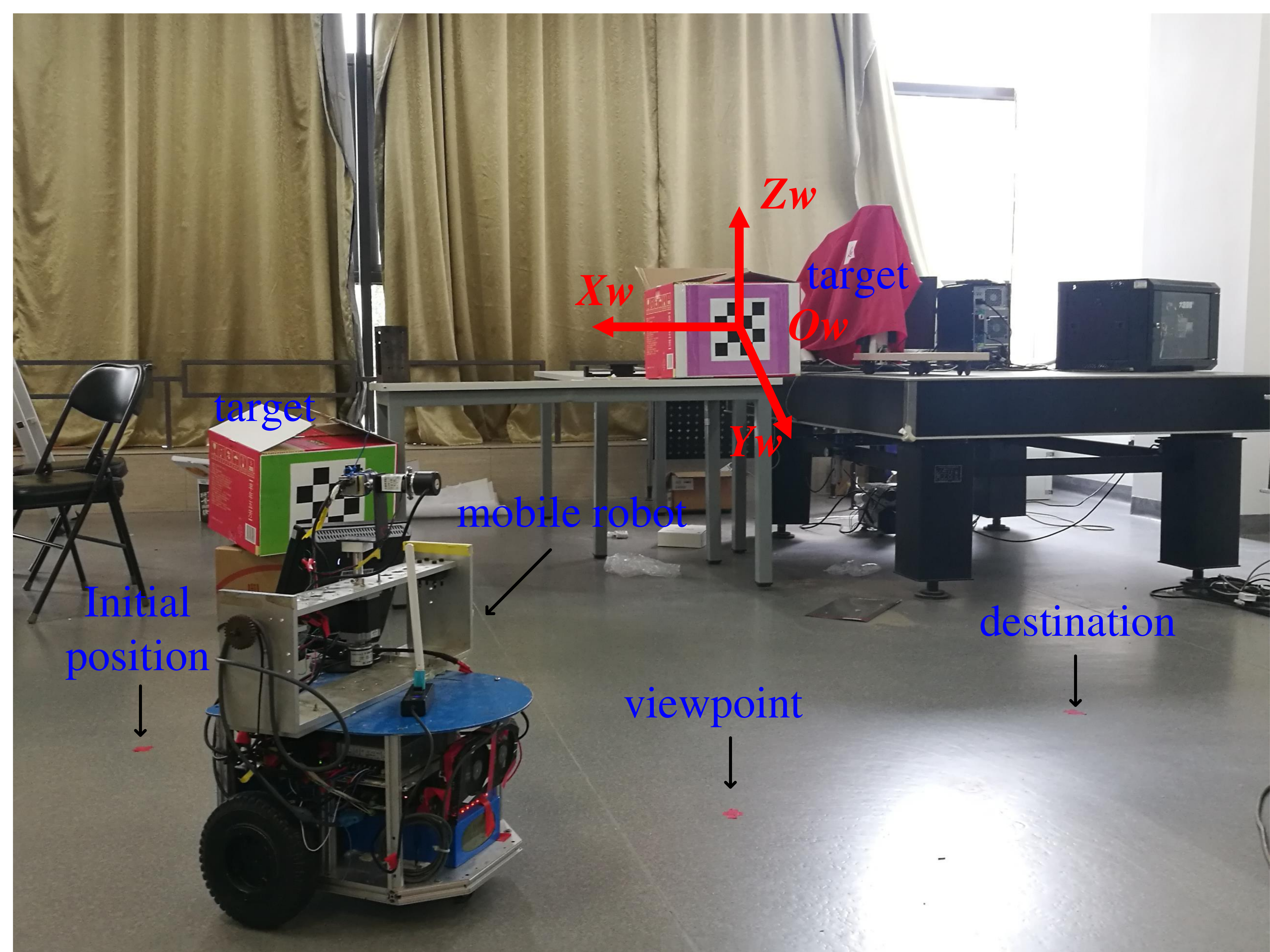}
\label{environment}
}
\subfigure[]{
\includegraphics[width=0.9in]{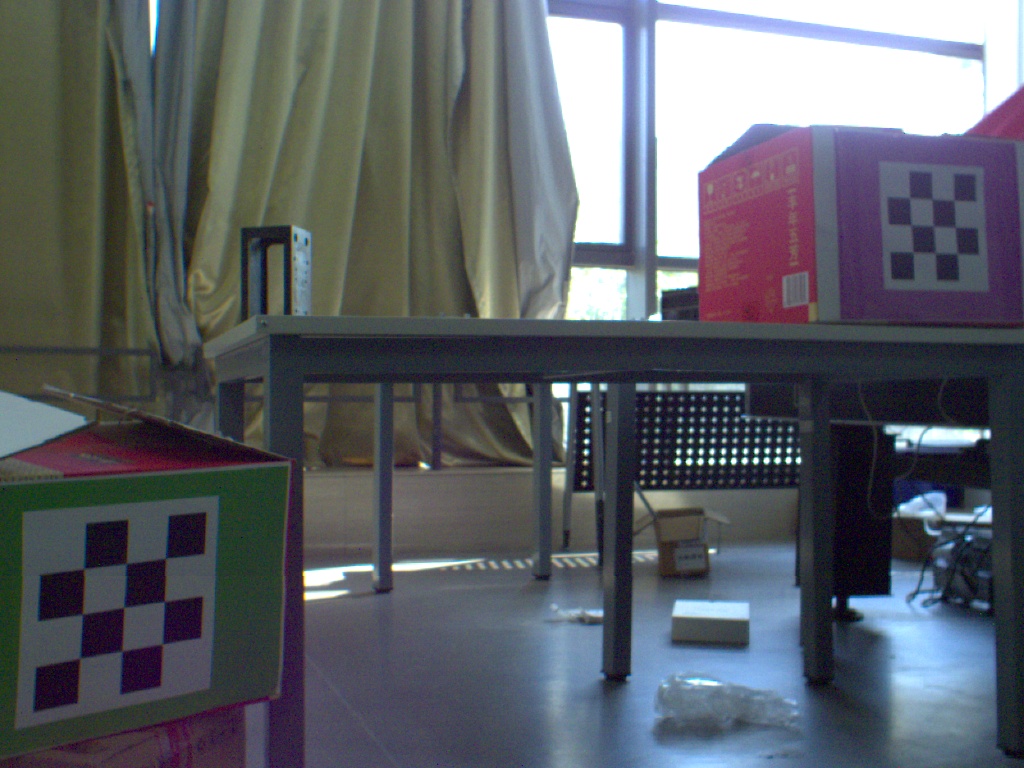}
\label{view2}
}
\subfigure[]{
\includegraphics[width=0.9in]{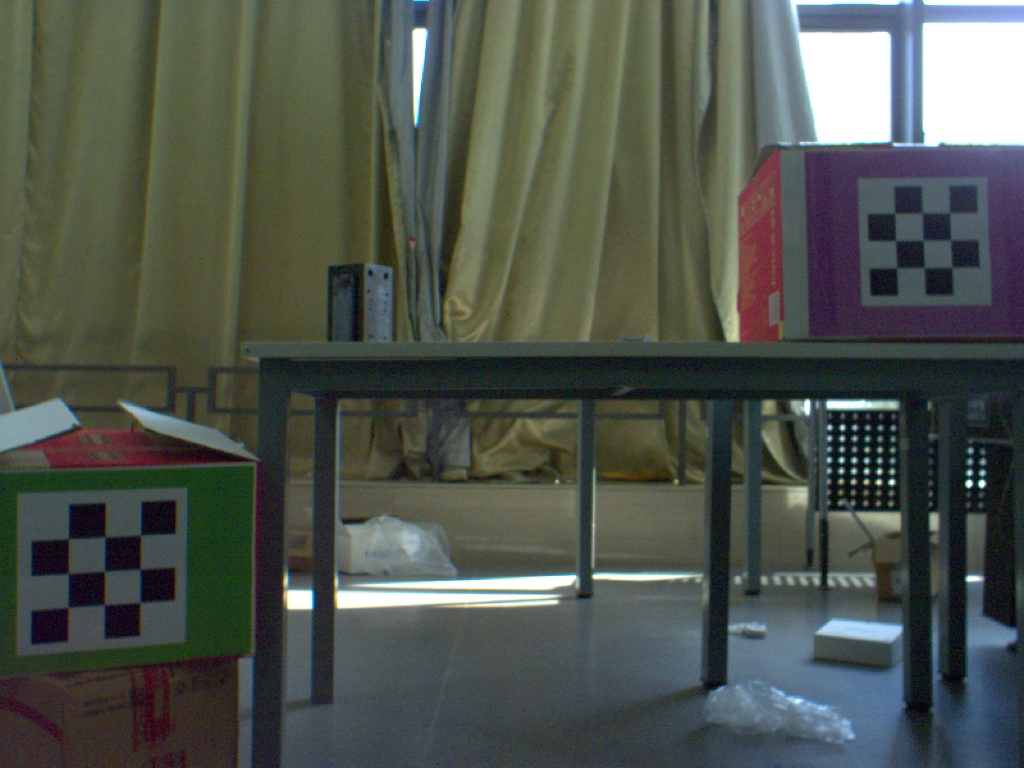}
\label{view3}
}
\setlength{\belowdisplayskip}{3pt}
\setlength{\belowdisplayskip}{3pt}
\setlength{\belowdisplayskip}{5pt}
\setlength{\belowdisplayskip}{3pt}
\subfigure[]{
\includegraphics[width=0.9in]{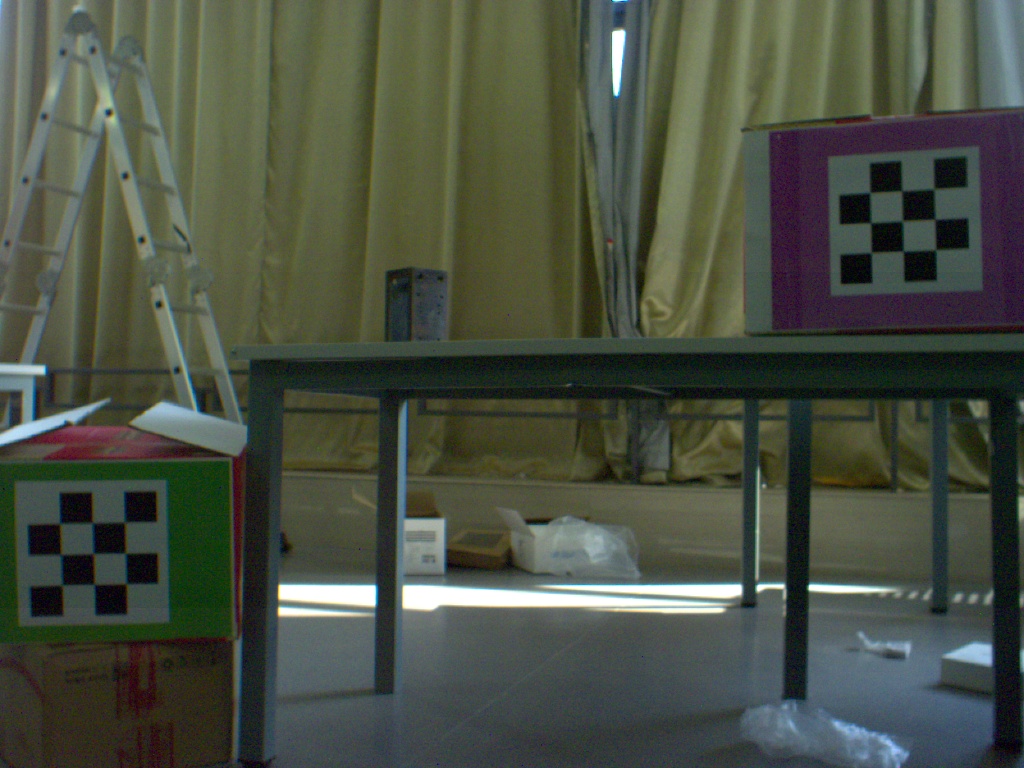}
\label{view4}
}
\setlength{\belowdisplayskip}{3pt}
\setlength{\belowdisplayskip}{3pt}
\setlength{\belowdisplayskip}{3pt}
\setlength{\belowdisplayskip}{3pt}
\subfigure[]{
\includegraphics[width=0.9in]{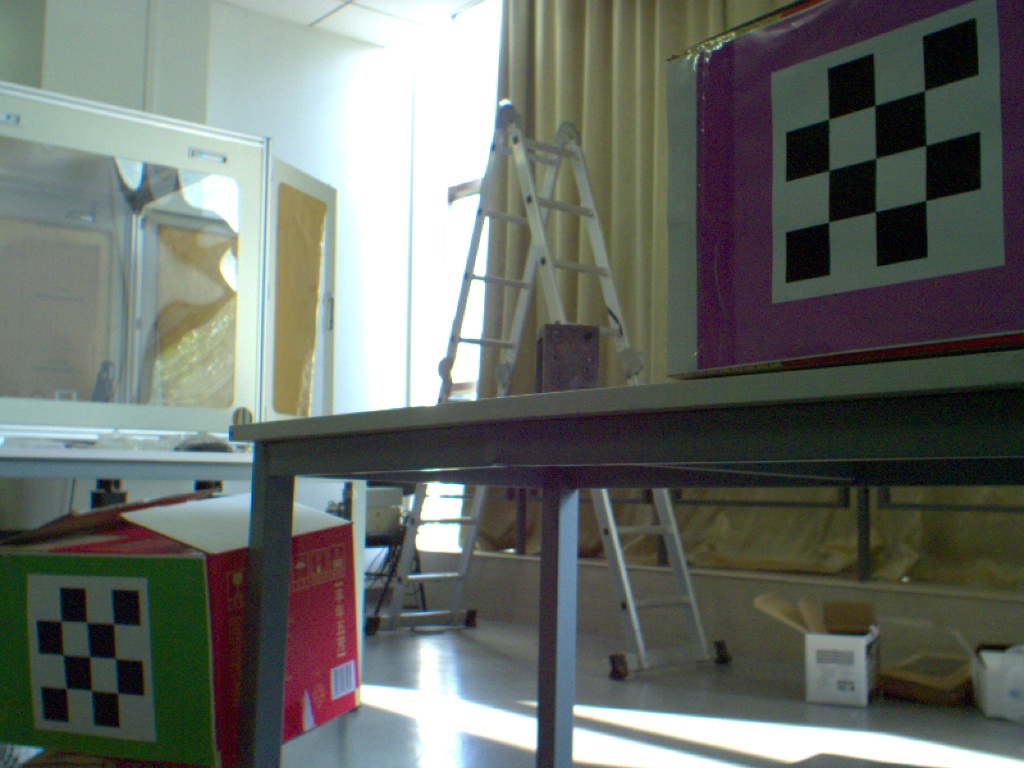}
\label{view5}
}
\caption{(a) The experimental scene. (b-f) The images taken by camera at position 1-4. (g) The trajectory of the mobile robot when it moves along the edge of the $RNA$. (h) The pixel distance when the mobile robot moves.}
\end{figure}

Besides simulations, experiments are conducted on a mobile robot equipped with a pan-tilt camera. The experimental scene is shown in Fig .\ref{environment}. The Horizontal-FOV angular aperture $\theta$ of camera is 1.1 rad, and the Vertical-FOV angular aperture $\varphi$ of camera is 0.85 rad. However, in order to increase some tolerance, the $\theta$ and $\varphi$ are set to 0.900 rad and 0.750 rad respectively, so $RNA$ calculated by our method becomes bigger than it really is.

In this experiment the mobile robot moves along the edge of $RNA$ as shown in Fig .\ref{TestTrajectory}. The initial position is (2.097, 1.671, 0), and the destination is (-0.9381, 1.0082, 0). In Fig .\ref{TestPixdis}, all distance values are positive, which means that under the guidance of $RNA$ all points can be kept in the camera's view during the motion of the mobile robot.
The images taken at positions 1-4 in Fig.\ref{TestTrajectory} correspond to Fig .\ref{view2}- \ref{view4}. They are all close to the edge of the $RNH$, and it can be seen that the right-most and left-most points are close to the edge of the image. Position 4 is also close to $RNV$, so it can be seen that the top-most and bottom-most points are close to the edge of the image. Combined with these images collected in different places, it can prove that true images are the same as what we predicted based on $RNA$.

\section{Conclusion}
In this paper, we propose a method  to calculate the $FOV$ constraint region. Based on the $FOV$ model of camera, two rectangle boxes $BH$ and $BV$ can be obtained. Then according to $BH$ and $BV$, the shape of $RNH$ and $RNV$ is derived. When the mobile robot is close to $RNH$, feature points will be close to the edge of image in the horizontal direction. When the mobile robot is close to $RNV$, feature points will be close to the edge of image in the vertical direction. By simulations and experiments, $RNA$ is proven to be close to the real $FOV$ constraint region. And $RNA$ can provide a good guidance for the mobile robot with pan-tilt camera to avoid these regions where the camera is unable to keep targets in a view.


\begin{thebibliography}{99}
\bibitem{Azouaoui2014Passively}
O.~Azouaoui and S.~B. T.~F. Salhi, ``Passively safe partial motion planning for
  mobile robots with limited field-of-views in unknown dynamic environments,''
  in \emph{IEEE International Conference on Robotics $\&$ Automation}, 2014.

\bibitem{Bouraine2012Provably}
S.~Bouraine, T.~Fraichard, and H.~Salhi, ``Provably safe navigation for mobile
  robots with limited field-of-views in dynamic environments,''
  \emph{Autonomous Robots}, vol.~32, no.~3, pp. 267--283, 2012.

\bibitem{Laurent2013Keeping}
L.~Burlion and H.~de~Plinval, ``Keeping a ground point in the camera field of
  view of a landing uav,'' in \emph{IEEE International Conference on Robotics
  $\&$ Automation}, 2013.

\bibitem{Lopez2007Switched}
G.~Lopez-Nicolas, S.~Bhattacharya, J.~J. Guerrero, C.~Sagues, and
  S.~Hutchinson, ``Switched homography-based visual control of differential
  drive vehicles with field-of-view constraints,'' in \emph{IEEE International
  Conference on Robotics $\&$ Automation}, 2007.

\bibitem{kang2019adaptive}
Z.~Kang, W.~Zou, H.~Ma, and Z.~Zhu, ``Adaptive trajectory tracking of wheeled
  mobile robots based on a fish-eye camera,'' \emph{International Journal of
  Control, Automation and Systems}, vol.~17, no.~9, pp. 2297--2309, 2019.

\bibitem{shao2013motion}
W.~Shao, Z.~J. Shen, L.~Meng, and S.~L. Sui, ``Motion constraint of planetary
  lander with overlapping field of view,'' in \emph{Proceedings of the 32nd
  Chinese Control Conference}.\hskip 1em plus 0.5em minus 0.4em\relax IEEE,
  2013, pp. 5159--5163.

\bibitem{choi2019deep}
J.~Choi, K.~Park, M.~Kim, and S.~Seok, ``Deep reinforcement learning of
  navigation in a complex and crowded environment with a limited field of
  view,'' in \emph{2019 International Conference on Robotics and Automation}.\hskip 1em plus 0.5em minus 0.4em\relax 2019, pp. 5993--6000.

\bibitem{Fang2012Adaptive}
Y.~Fang, X.~Liu, and X.~Zhang, ``Adaptive active visual servoing of
  nonholonomic mobile robots,'' \emph{IEEE Transactions on Industrial
  Electronics}, vol.~59, no.~1, pp. 486--497, 2012.

\bibitem{stolkin2008efficient}
R.~Stolkin, I.~Florescu, M.~Baron, C.~Harrier, and B.~Kocherov, ``Efficient
  visual servoing with the abcshift tracking algorithm,'' in \emph{2008 IEEE
  International Conference on Robotics and Automation}.\hskip 1em plus 0.5em
  minus 0.4em\relax 2008, pp. 3219--3224.
\bibitem{lopez2017formation}
G.~L{\'o}pez-Nicol{\'a}s, M.~Aranda, and Y.~Mezouar, ``Formation of
  differential-drive vehicles with field-of-view constraints for enclosing a
  moving target,'' in \emph{2017 IEEE International Conference on Robotics and
  Automation}.\hskip 1em plus 0.5em minus 0.4em\relax 2017, pp.
  261--266.
\bibitem{Park2012Novel}
D.~H. Park, J.~H. Kwon, and I.~J. Ha, ``Novel position-based visual servoing
  approach to robust global stability under field-of-view constraint,''
  \emph{IEEE Transactions on Industrial Electronics}, vol.~59, no.~12, pp.
  4735--4752, 2012.

\bibitem{zhu2019scalable}
Z.~Zhu and W.~Zou, ``Scalable and occlusion-aware multi-cues correlation filter
  for robust stereo visual tracking,'' \emph{International Journal of Robotics
  and Automation}, vol.~34, no.~5.
\bibitem{Fan2018Robust}
K.~Fan, Z.~Li, C.~Yang, K.~Fan, Z.~Li, and C.~Yang, ``Robust tube-based
  predictive control for visual servoing of constrained differential-drive
  mobile robots,'' \emph{IEEE Transactions on Industrial Electronics}, vol.~PP,
  no.~99, pp. 1--1, 2018.

\bibitem{Le2016Optimal}
V.~C. Le and T.~C. Pham, ``Optimal tracking a moving target for integrated
  mobile robot-pan tilt-stereo camera,'' in \emph{IEEE International Conference
  on Advanced Intelligent Mechatronics}, 2016.

\bibitem{Bhattacharya2005Path}
S.~Bhattacharya, R.~Murrieta-Cid, and S.~Hutchinson, ``Path planning for a
  differential drive robot: minimal length paths - a geometric approach,'' in
  \emph{IEEE/RSJ International Conference on Intelligent Robots $\&$ Systems},
  2005.

\bibitem{Bhattacharya2006Controllability}
S.~Bhattacharya and S.~Hutchinson, ``Controllability and properties of optimal
  paths for a differential drive robot with field-of-view,'' in \emph{IEEE
  International Conference on Robotics $\&$ Automation}, 2006.

\bibitem{Bhattacharya2007Optimal}
S.~Bhattacharya, R.~Murrieta-Cid, and S.~Hutchinson, ``Optimal paths for
  landmark-based navigation by differential-drive vehicles with field-of-view
  constraints,'' \emph{IEEE Transactions on Robotics}, vol.~23, no.~1, pp.
  47--59, 2007.

\bibitem{Salaris2010Shortest}
P.~Salaris, D.~Fontanelli, L.~Pallottino, and A.~Bicchi, ``Shortest paths for a robot with nonholonomic and field-of-view constraints,'' \emph{IEEE
  Transactions on Robotics}, vol.~26, no.~2, pp. 269--281, 2010.

\bibitem{Salaris2013Shortest}
P.~Salaris, A.~Cristofaro, L.~Pallottino, and A.~Bicchi, ``Shortest paths for
  wheeled robots with limited field-of-view: Introducing the vertical
  constraint,'' in \emph{Decision and Control}, 2013, pp. 5143--5149.

\bibitem{Salaris2011From}
P.~Salaris, L.~Pallottino, S.~Hutchinson, and A.~Bicchi, ``From optimal
  planning to visual servoing with limited fov,'' in \emph{IEEE/RSJ
  International Conference on Intelligent Robots $\&$ Systems}, 2011.

\bibitem{Salaris2015Epsilon}
P.~Salaris, A.~Cristofaro, L.~Pallottino, and A.~Bicchi, ``Epsilon-optimal
  synthesis for vehicles with vertically bounded field-of-view,'' \emph{IEEE
  Transactions on Automatic Control}, vol.~60, no.~5, pp. 1204--1218, 2015.




\end{thebibliography}
\end{document}